\documentclass[sn-apa,iicol]{sn-jnl}

\usepackage{graphicx}%
\usepackage{multirow}%
\usepackage{amsmath,amssymb,amsfonts}%
\usepackage{amsthm}%
\usepackage{mathrsfs}%
\usepackage[title]{appendix}%
\usepackage[dvipsnames]{xcolor}
\usepackage{textcomp}%
\usepackage{manyfoot}%
\usepackage{booktabs}%
\usepackage{algorithm}%
\usepackage{algorithmicx}%
\usepackage{algpseudocode}%
\usepackage{listings}%

\usepackage{subcaption}
\usepackage{array, tabularx, bm}
\usepackage{adjustbox}
\usepackage{arydshln}
\usepackage{pifont}
\usepackage{capt-of}
\usepackage{float,wrapfig}
\usepackage{comment}
\usepackage{url}
\usepackage{mathtools, empheq}
\usepackage{enumitem}
\usepackage{colortbl}

\newcommand{\colearnplus}{Co-learn\texttt{++}}
\newcommand{\plus}{\texttt{++}}
\newcommand{\cmark}{\ding{51}} 
\newcommand{\xmark}{\ding{55}} 

\newcolumntype{P}[1]{>{\centering\arraybackslash}p{#1}}

\algnewcommand\algorithmicforeach{\textbf{for each}}
\algdef{S}[FOR]{ForEach}[1]{\algorithmicforeach\ #1\ \algorithmicdo}

\newcommand{\green}[1]{{\color{green} #1}}
\newcommand{\red}[1]{{\color{red} #1}}

\begin{document}

\title[Source-Free Domain Adaptation Guided by Vision and Vision-Language Pre-Training]{\centering Source-Free Domain Adaptation \\ Guided by Vision and Vision-Language Pre-Training}


\author*[1]{\fnm{Wenyu} \sur{Zhang}}\email{zhang\_wenyu@i2r.a-star.edu.sg}

\author[1]{\fnm{Li} \sur{Shen}}\email{lshen@i2r.a-star.edu.sg}

\author[1,2]{\fnm{Chuan-Sheng} \sur{Foo}}\email{foo\_chuan\_sheng@i2r.a-star.edu.sg}

\affil[1]{\orgname{Institute for Infocomm Research (I$^\text{2}$R), Agency for Science, Technology and Research (A*STAR)}}

\affil[2]{\orgname{Centre for Frontier AI Research (CFAR), Agency for Science, Technology and Research (A*STAR)}}

\abstract{Source-free domain adaptation (SFDA) aims to adapt a source model trained on a fully-labeled source domain to a related but unlabeled target domain. While the source model is a key avenue for acquiring target pseudolabels, the generated pseudolabels may exhibit source bias. In the conventional SFDA pipeline, a large data (e.g. ImageNet) pre-trained feature extractor is used to initialize the source model at the start of source training, and subsequently discarded. Despite having diverse features important for generalization, the pre-trained feature extractor can overfit to the source data distribution during source training and forget relevant target domain knowledge. Rather than discarding this valuable knowledge, we introduce an integrated framework to incorporate pre-trained networks into the target adaptation process. The proposed framework is flexible and allows us to plug modern pre-trained networks into the adaptation process to leverage their stronger representation learning capabilities.
For adaptation, we propose the \textit{Co-learn} algorithm to improve target pseudolabel quality collaboratively through the source model and a pre-trained feature extractor. Building on the recent success of the vision-language model CLIP in zero-shot image recognition, we present an extension \textit{\colearnplus} to further incorporate CLIP's zero-shot classification decisions.
We evaluate on 4 benchmark datasets and include more challenging scenarios such as open-set, partial-set and open-partial SFDA. Experimental results demonstrate that our proposed strategy improves adaptation performance and can be successfully integrated with existing SFDA methods. Project code is available at \url{https://github.com/zwenyu/colearn-plus}.}

\keywords{source-free domain adaptation, pseudolabeling, pre-trained networks}

\maketitle

\section{Introduction}
\label{sec: introduction}

Deep neural networks have demonstrated remarkable proficiency across a spectrum of applications, but their effectiveness typically relies on the assumption that training \emph{(source domain)} and test \emph{(target domain)} data distributions are the same. This assumption can be violated in practice when the source data does not fully represent the entire data distribution due to difficulties of real-world data collection. Target samples distributed differently from source samples (due to factors such as background, illumination and style variations~\citep{gulrajani2020domainbed,wilds}) manifest as instances of \emph{domain shift} (also known as \emph{covariate shift}), and can severely degrade model performance.

\emph{Domain adaptation} (DA) aims to address the challenge of domain shift by transferring knowledge from a fully-labeled source domain to a related but unlabeled target domain. The classic setting of \emph{unsupervised domain adaptation} assumes both source and target data are jointly available for training~\citep{wilson2020dasurvey}. Motivated by the theoretical bound on target risk derived in \citep{BenDavid2010domainadaptation}, a fundamental strategy is to minimize the discrepancy between source and target features to learn domain-invariant features~\citep{Ganin2016DANN,Li2018CDANN,li2018mmd,Sun2016DeepCORAL, zhang2019mdd,hu2020gsda,kang2019can,gu2020rsda,Xu2019sfan}. However, access to source data can be impractical due to data privacy concerns. Recently, the \emph{source-free domain adaptation} (SFDA) setting is introduced~\citep{liang2020shot} to divide unsupervised domain adaptation into two stages: (i) source training stage: training network with fully labeled source data, and (ii) target adaptation stage: adapting source model with unlabeled target data. An example use case in a corporate context is when a vendor has collected data to train a source model, and clients seek to address a similar task in their own environments. However, data sharing for joint training is precluded due to proprietary or privacy considerations.
In this use case, the vendor makes available the source model, and allows clients to adapt it with their available resources. Our focus in this work is to assume the role of the clients.

\begin{figure}
  \centering
  \begin{subfigure}{0.48\linewidth}
    \includegraphics[width=\linewidth]{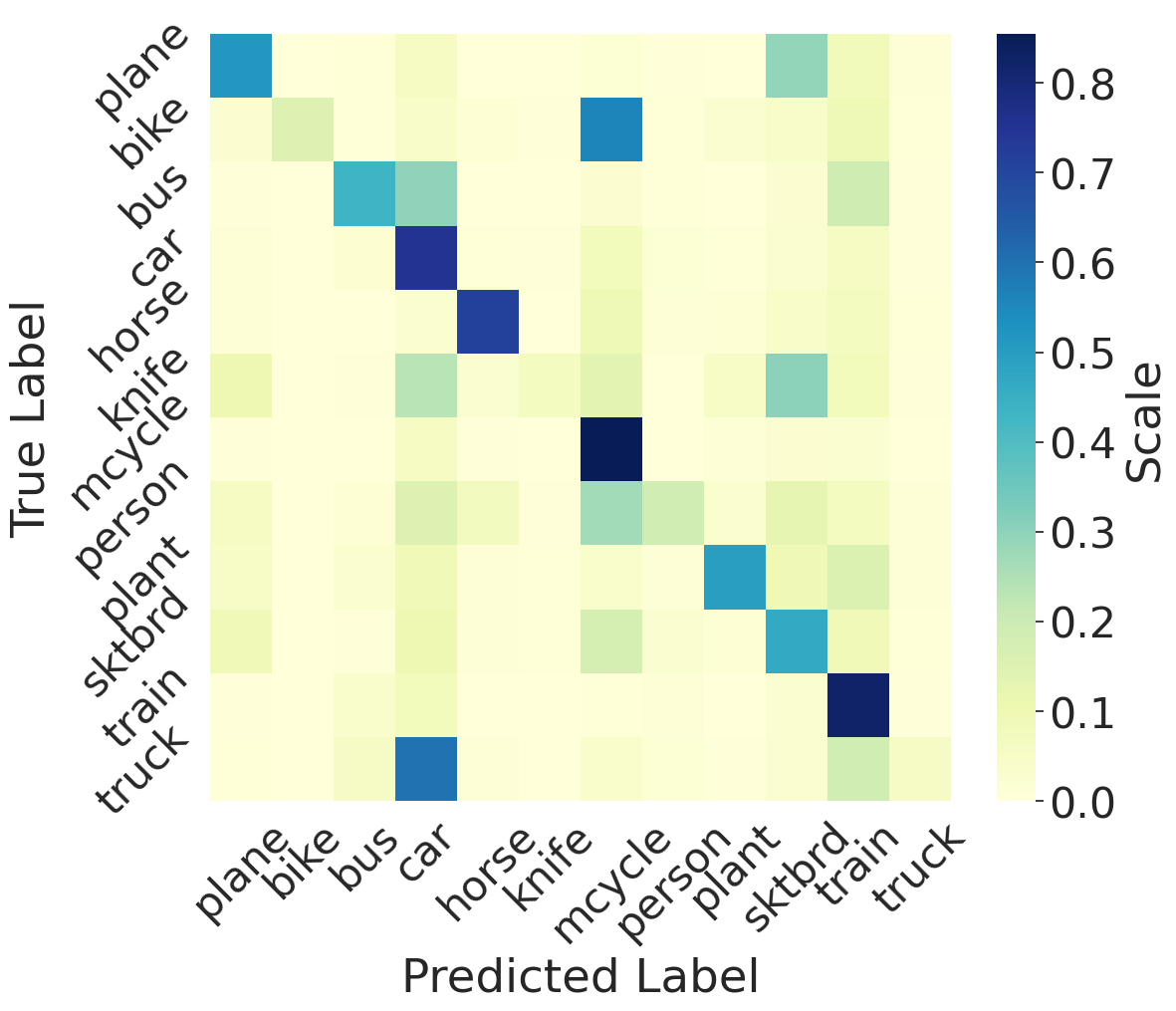}
    \caption{Confusion matrix of true vs. predicted labels}
    \label{fig: visda_confusion_matrix}
  \end{subfigure}
  \hfill
  \begin{subfigure}{0.48\linewidth}
    \includegraphics[width=\linewidth]{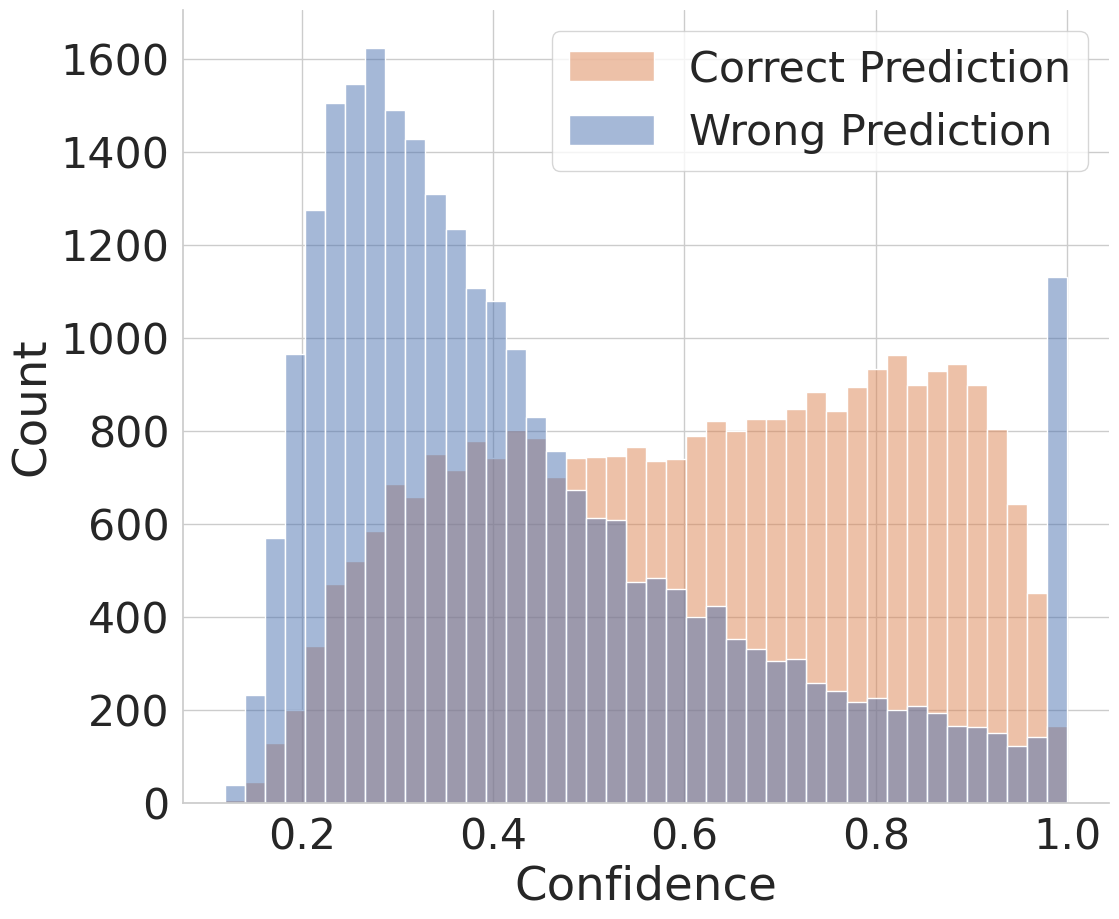}
    \caption{Distribution of prediction confidence}
    \label{fig: visda_confidence}
  \end{subfigure}
  
  \caption{VisDA-C source-trained ResNet-101 produces unreliable pseudolabels on target samples, and is over-confident on a significant number of incorrect predictions.}
  \label{fig: visda_prediction}
\end{figure}

The typical approach to obtain the source model is to train a selected pre-trained network on source data with supervised loss.  
For adaptation, existing SFDA methods generate or estimate source-like representations to align source and target distributions~\citep{qiu2021prototypegen,ding2022sfdade}, exploit local clustering structures in the target data~\citep{yang2021gsfda,Yang2021ExploitingTI,Yang2022AttractingAD}, and learn semantics through self-supervised tasks~\citep{xia2021a2net,liang2021shotplus}.
\citep{liang2020shot,Kundu2020TowardsIM,li20203cgan} use the source model to generate target pseudolabels for finetuning, and \citep{kim2021sfda,chen2021noisy,liang2021shotplus} further select samples based on low-entropy or low-loss criterion. However, model calibration is known to degrade under distribution shift~\citep{ovadia2019uncertainty}. We observe in Figure~\ref{fig: visda_prediction} that target pseudolabels produced by the source model can be biased, and outputs such as prediction confidence (and consequently entropy and loss) may not reflect accuracy and hence cannot reliably be used alone to improve pseudolabel quality.

\begin{figure}[tb]
    \centering
    \includegraphics[width=\linewidth]{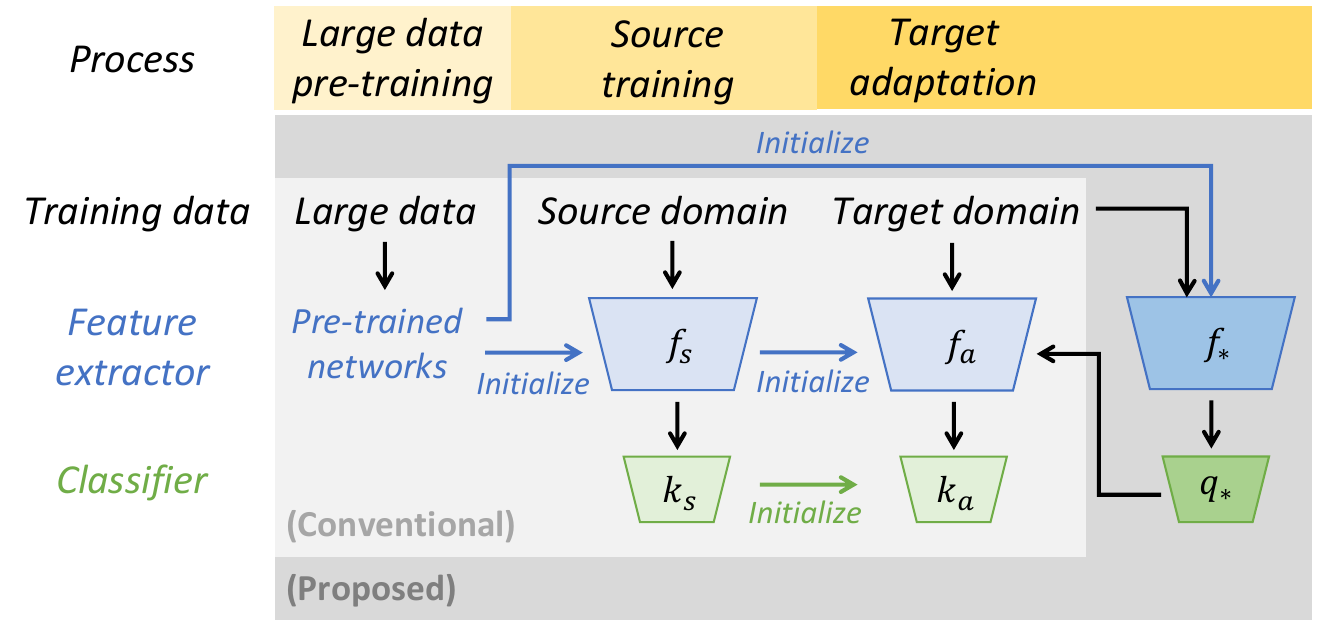}
    
    \caption{Overview of conventional and proposed framework. We propose incorporating pre-trained networks during target adaptation. For the pre-trained network, we can plug in the same network used for source model initialization, or a different network that potentially has better feature extraction capabilities on the target domain.}
    \label{fig: overview_setting}
\end{figure}

In the conventional SFDA pipeline, pre-trained networks are utilized solely for source model initialization. We reconsider the appropriateness of this role in domain shift settings. 
From the SFDA pipeline illustrated in Figure~\ref{fig: overview_setting}, large data pre-trained weights such as ImageNet weights are conventionally used to initialize source models and subsequently discarded after source training. While the large data pre-trained model initially has diverse features important for generalization~\citep{chen2021contrastive}, finetuning on source data can cause it to overfit to source distribution and forget pre-existing target information. Relying on the resulting biased source model to guide target adaptation (e.g. through generation of target pseudolabels) risks the target model inheriting the source bias.
We seek to answer the questions:
\begin{itemize}[noitemsep]
    \item \emph{Can finetuning a pre-trained network on source data cause it to forget relevant target domain knowledge?}
    \item \emph{Given a source model, whether and how a pre-trained network can help its adaptation?}
\end{itemize}

Discarding pre-trained networks directly after source training risks simultaneously discarding any relevant target domain knowledge they may hold.
We propose to integrate these pre-trained networks into the target adaptation process, as depicted in Figure~\ref{fig: overview_setting}, to provide a viable channel to distil useful target domain knowledge from them after the source training stage. Regarding the pre-trained network to be integrated, we have the option to select the one employed for source model initialization or select another network that may have superior feature extraction capabilities on the target domain, thereby harnessing the respective advantages in correcting source model bias: 
\begin{itemize}[noitemsep]
    \item Restoration of target domain knowledge lost from the pre-trained network during source training;
    \item Insertion of target domain knowledge from a more powerful pre-trained network into the source model.
\end{itemize}
For instance, modern foundation models enjoy better generalizability following the development of new architectures, extensive training data and customized training schemes, and can provide positive guidance during target adaptation.

\begin{figure}[h!]
    \centering
    \begin{subfigure}{0.95\columnwidth}
        \centering
        \includegraphics[width=\linewidth]{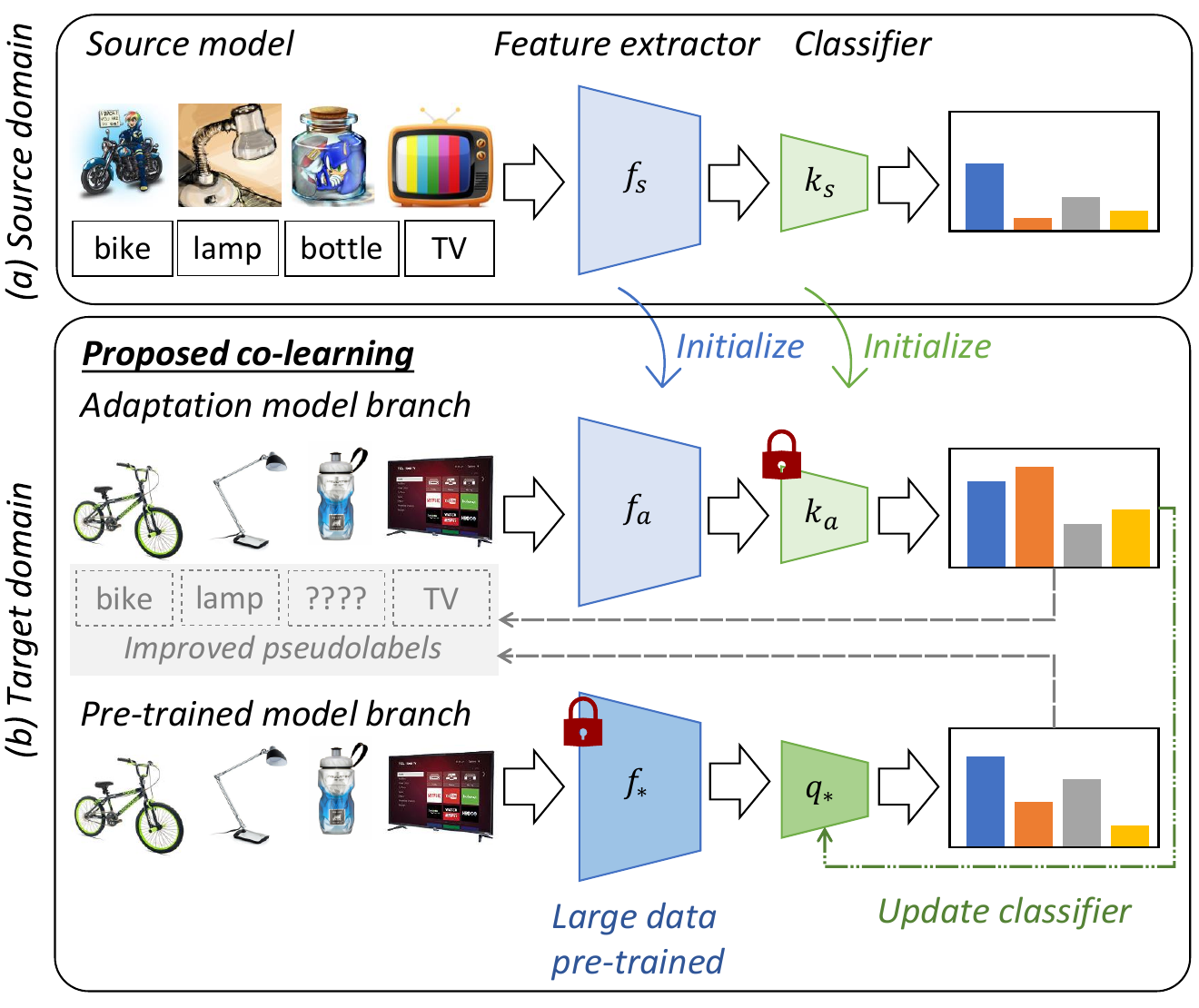}
        \caption{Co-learning framework} 
        \label{fig: overview framework}
    \end{subfigure}

    \begin{subfigure}{0.95\columnwidth}
        \centering
        \includegraphics[width=\linewidth]{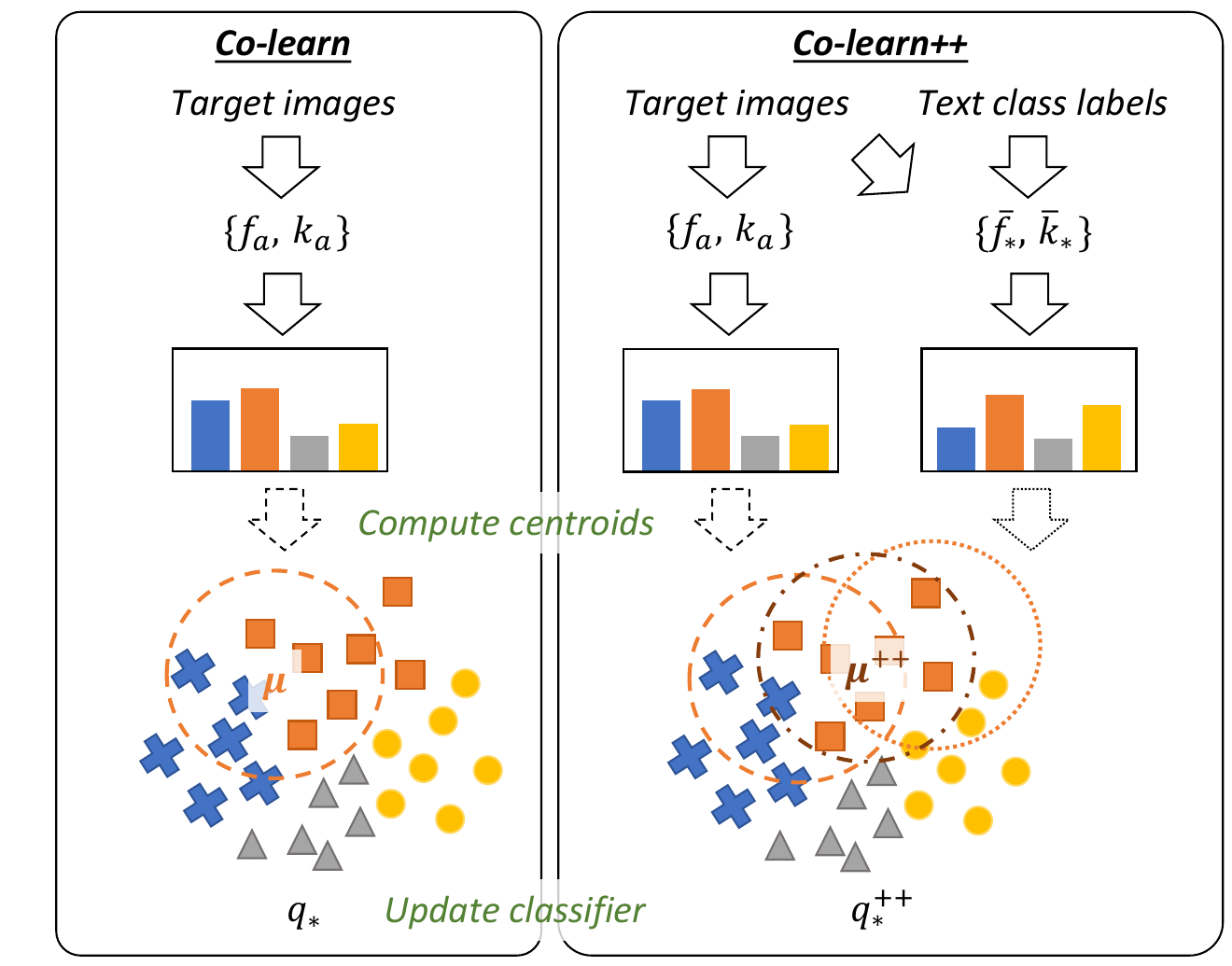}
        \caption{Estimation of task-specific classification head in pre-trained model branch} 
        \label{fig: overview classifier}
    \end{subfigure}
    
    \caption{Overview of proposed strategy: (i) Source model trained on source domain is provided. (ii) We adapt the source model through a co-learning strategy where the adaptation model $\{f_a, k_a\}$ and a large data pre-trained feature extractor $f_*$ collectively produce more reliable pseudolabels for finetuning. To estimate a new classification head on $f_*$, the Co-learn algorithm computes a nearest-centroid-classifier $q_*$ weighted by adaptation model predictions, and the \colearnplus\ algorithm additionally integrates CLIP zero-shot predictions from the text-based classifier $\{\bar{f}_a, \bar{k}_a\}$ to compute $q^{\plus}_*$.}
    \label{fig: overview}
\end{figure}

To facilitate effective knowledge transfer, we design a simple two-branch co-learning strategy where the adaptation and pre-trained model branch iteratively updates to collaboratively generate more accurate target pseudolabels. The strategy is agnostic to network architecture, and the resulting pseudolabels can be readily applied to existing SFDA methods as well. We provide an overview of our strategy in Figure~\ref{fig: overview}. The `Co-learn' algorithm focuses on integrating a pre-trained vision encoder (e.g. from a ImageNet model), and was published in our prior work \citep{zhang2023colearn}. One limitation of the approach is that it relies on pseudolabels derived from the source model, which can be biased, to fit a task-specific classifier on the pre-trained envison encoder. Motivated by the recent success of the pre-trained vision-language model CLIP \citep{Radford2021clip} in zero-shot image recognition, we provide an extension `\colearnplus' to integrate the CLIP vision encoder for co-learning and to refine the fitted task-specific classifier with zero-shot classification decisions from CLIP's text-based classifier. The co-learning approach is effective in integrating knowledge from target-adapted CLIP as well. All methodology and experiments with CLIP are new materials extending our prior work. We also include additional experiments and analysis on the effectiveness of the proposed framework and strategy. Our contributions are summarized as follows:
\begin{itemize}[noitemsep]
    \item We observe that finetuning pre-trained networks on source data can cause overfitting and loss of generalizability for the target domain;
    \item Based on the above observation, we propose an integrated SFDA framework to incorporate pre-trained networks into the target adaptation process;
    \item We propose a simple co-learning strategy to distil useful target domain information from a pre-trained network to improve target pseudolabel quality;
    \item The `Co-learn' algorithm focuses on integrating a pre-trained vision encoder, and the `\colearnplus' algorithm integrates the CLIP vision encoder and refines the fitted classifier with zero-shot classification decisions from CLIP's text-based classifier;
    \item We evaluate on 4 benchmark SFDA image classification datasets, and also evaluate on more challenging SFDA scenarios such as open-set, partial-set and open-partial SFDA;
    \item We demonstrate performance improvements by the proposed framework and strategy (including just reusing the pre-trained network in source model initialization) and by incorporating the proposed strategy in existing SFDA methods.
\end{itemize}

\section{Related Works}
\label{sec: related works}

\subsection{Unsupervised domain adaptation}

In traditional unsupervised DA, networks are trained jointly on labeled source and unlabeled target dataset to optimize task performance on the target domain~\citep{wilson2020dasurvey}. A popular strategy is to learn domain-invariant features via minimizing domain discrepancy measured by a inter-domain distance or adversarial loss~\citep{Ganin2016DANN,Li2018CDANN,li2018mmd,Sun2016DeepCORAL, zhang2019mdd,hu2020gsda,kang2019can,gu2020rsda,Xu2019sfan}. \citep{zhao2021madan} considers pixel-level alignment. \citep{lu2020star} learns a distribution of classifiers to better identify local regions of misalignment between source and target domains, and \citep{cui2020gvb,na2021fixbi} facilitate alignment between distant domains by generating intermediate domains.
\citep{Li2021TSA} augments source data to assume target style, and \citep{kumar2023freq} applies frequency transformation to reduce the domain gap.
Other methods encourage target representations to be more class-discriminative by learning cluster structure~\citep{tang2020srdc} or minimizing uncertainty~\citep{jin2020mcc}. Amongst confidence-based methods, \citep{gu2020rsda} trains a spherical neural network to select target samples for pseudolabeling, \citep{na2021fixbi,french2018se} selects high confidence predictions as positive pseudolabels and \citep{french2018se} applies mean teacher from semi-supervised learning. These methods assume simultaneous access to both source and target data, and are not suitable for SFDA where source data is not available for joint training.

\subsection{Source-free domain adaptation}
\label{subsec: source-free domain adaptation}

In SFDA, the source model is adapted with an unlabeled target dataset. \citep{kundu2022concurrent,kundu2022balancing} train on the source domain with auxiliary tasks or augmentations and \citep{roy2022usfan} calibrates uncertainties with source data to obtain improved source models, but these strategies cannot be applied on the client-side where the source model is fixed and source data is not accessible.
Some methods exploit local clustering structures in the target dataset to learn class-discriminative features~\citep{yang2021gsfda,Yang2021ExploitingTI,Yang2022AttractingAD}, and learn semantics through self-supervised tasks such as rotation prediction~\citep{xia2021a2net,liang2021shotplus}. \citep{sanqing2022BMD} proposes multi-centric clustering, but it is not straightforward how to select the cluster number for each dataset. \citep{xia2021a2net} learns a new target-compatible classifier, and \citep{li20203cgan} generates target-style data with a generative adversarial network to improve predictions.
Other methods generates source-like representations~\citep{qiu2021prototypegen} or estimates source feature distribution to align source and target domains~\citep{ding2022sfdade}. \citep{luo2024crots} generates diverse samples and \citep{tang2024search} uses intermediate proxy distributions to bridge large domain gaps. \citep{dong2021caida} and \citep{jin2023seprepnet} leverage multiple source domains. \citep{liang2020shot,Kundu2020TowardsIM,li20203cgan} use the source model to generate pseudolabels for the target dataset, and align target samples to the source hypothesis through entropy minimization and information maximization. To improve pseudolabel quality, \citep{kim2021sfda,chen2021noisy,liang2021shotplus} select target samples with low entropy or loss for pseudolabeling and finetuning the source model.
We find that source model outputs may not reliably reflect target pseudolabel accuracy due to domain shift in Figure~\ref{fig: visda_prediction}. Instead of relying solely on possibly biased pseudolabels produced by the source model as in existing works, we make use of a pre-trained network to rectify the source bias and produce more reliable pseudolabels. 

Recent works leverage the strong image recognition capabilities of CLIP for source-free domain adaptation. POUF~\citep{tanwisuth2023pouf} and ReCLIP~\citep{xuefeng2023reclip} directly adapt CLIP models on the target domain, resulting in large target models (depending on the choice of backbone) which may not be suitable for light-weight applications. DALL-V~\citep{Zara2023dallv} requires source pre-training and target adaptation on CLIP models, and is not suitable in scenarios where the source model is provided as is. Moreover, DALL-V is a work on video domain adaptation while we focus on images. Our proposed approach does not require custom source pre-training, and results in a target model that performs competitively or better than the larger CLIP model it co-learned with.

\section{Role of Pre-trained Networks in SFDA}
\label{sec: role of pre-trained networks}

\subsection{Conventional role}

In the conventional source-free domain adaptation framework in Figure~\ref{fig: overview_setting}, during source training, the source model is initialized with pre-trained weights and then trained on source data with supervised objective. The pre-trained weights are learned from large and diverse datasets such as ImageNet (IN). Warm-starting the training process with pre-trained weights is a common practice in computer vision to improve the generalization capabilities of the trained model and to mitigate the requirement for substantial quantities of training data.

\subsection{Considerations on target generalizability}

In SFDA, the goal is to accurately estimate the conditional probability $p(y_t|x_t)$ for target input $x_t \in \mathcal{X}_t$ and output $y_t \in \mathcal{Y}_t$. For a model with feature extractor (FE) $f$ and classifier (CLF) $k$ and normalizing function $\sigma$, the estimated probabilities are $\left[\hat{p}(y|x_t)\right]_{y\in\mathcal{Y}_t} = \sigma\left(k(f(x_t))\right)$. Since the source model is trained to maximize accuracy metrics on the source data, it may not be the network that maximizes the accuracy of target estimates $\hat{p}(y_t|x_t)$. Large data pre-trained networks, such as the ones used as initializations for source training or powerful foundation models, may be more compatible with the target domain instead. That is, firstly, class-discriminative information useful for the target domain may be lost from the pre-trained network during source training as the source model learns to fit exclusively to the source data distribution. Secondly, the information extracted up to the source training stage may be inadequate for proficient image recognition on the target domain (e.g. due to network architecture design), such that we need to borrow strength from more sophisticated information extraction techniques. We list the mapping from input to feature space, and from feature to output space for the pre-trained model and source model below. We take ImageNet (IN) as the example for pre-training data, but the discussion can be similarly applied in the context of other large pre-training datasets.

\begin{footnotesize}
\begin{empheq}[box=\fbox]{alignat*=5}
    &\textbf{Input} &&\xrightarrow{\makebox[8mm] {\textbf{FE}}} &&\quad \textbf{Feature} &&\xrightarrow{\makebox[8mm] {\textbf{CLF}}} &&\quad \textbf{Output} \\
    &\textit{IN data} &&\xrightarrow{\makebox[8mm] {$f_*$}} &&\quad \textit{IN feature} &&\xrightarrow{\makebox[8mm] {$k_*$}} &&\quad \textit{IN class} \\    
    &\textit{source} &&\xrightarrow{\makebox[8mm] {$f_s$}} &&\quad \textit{source feature} &&\xrightarrow{\makebox[8mm] {$k_s$}} &&\quad \textit{source class}
\end{empheq}
\end{footnotesize}

The ImageNet and source model are optimized for the accuracy of ImageNet estimates $\hat{p}(y_*|x_*)$ and source domain estimates $\hat{p}(y_s|x_s)$, respectively. Their accuracy on target domain estimates $\hat{p}(y_t|x_t)$ depends on:
\begin{enumerate}[noitemsep]
    \item Similarity between training and target inputs (i.e. images);
    \item Robustness of input-to-feature mapping under distribution differences between training and target inputs (covariate shift);  
    \item Similarity between training and target outputs (i.e. class labels).
\end{enumerate}
Pre-trained models can be advantageous in terms of the above criteria because (1) the larger and more diverse pre-training dataset is not source-biased and may better capture the target input distribution, (2) modern state-of-the-art network architectures are designed to learn more robust input-to-feature mappings, enabling better transfer to target tasks, (3) recent vision-language models can leverage textual information in class labels to perform zero-shot recognition on new tasks with different label spaces. However, in general vision models, since the pre-training and target label space differ, the pre-trained classifier $k_*$ needs to be replaced. One advantage of the source model is that it is trained for the target label space, but it may suffer from a lack of generalization to different input distributions. 

\noindent\textbf{Examples of target generalizability of pre-trained networks.}
Firstly, in Table~\ref{tab: comparison_resnet}, an example where target domain class-discriminative information is lost after source training is the Clipart-to-Real World ($C\rightarrow R$) transfer in Office-Home dataset. The source ResNet-50 is initialized with ImageNet ResNet-50 weights at the start of source training. The target domain (i.e. Real World) is more distant in style to the source domain (i.e. Clipart) than to ImageNet, such that oracle target accuracy (computed by fitting a classification head on the feature extractor using fully-labeled target samples) drops from $86.0\%$ to $83.3\%$ after source training. 

\begin{table}[tb]
\centering
\setlength{\tabcolsep}{2pt}
\begin{adjustbox}{max width=\columnwidth}
\begin{tabular}{l*{6}{c}}
\toprule[1pt]\midrule[0.3pt]

\textbf{Feature extractor} & \multicolumn{2}{c}{$\mathbf{C\rightarrow R}$}  & \multicolumn{2}{c}{$\mathbf{A\rightarrow C}$}  &  \multicolumn{2}{c}{$\mathbf{R\rightarrow A}$} \\ \cmidrule(lr){2-3} \cmidrule(lr){4-5} \cmidrule(lr){6-7}
                    & Oracle    & Co-learn  & Oracle    & Co-learn  & Oracle    & Co-learn \\ \midrule
Source ResNet-50        & 83.3      & 74.1      & \textbf{69.5}      & \emph{53.9}      & \emph{81.8}      & 70.5 \\
ImageNet ResNet-50   & \emph{86.0}      & \emph{79.4}      & 65.5      & 51.8      & 81.2      & \emph{71.1} \\
ImageNet ResNet-101  & \textbf{87.0}      & \textbf{80.4}      & \emph{68.0}      & \textbf{54.6}      & \textbf{82.5}      & \textbf{72.4} \\
\midrule[0.3pt]\bottomrule[1pt]
\end{tabular}
\end{adjustbox}
\caption{Comparison of source versus ImageNet models on Office-Home. The `Oracle' column lists target accuracy obtained by source and ImageNet feature extractors + oracle classification head, demonstrating the presence of target information loss due to source training (e.g. in $C\rightarrow R$). The `Co-learn' column lists target accuracy of adapted ResNet-50 co-learned with source and ImageNet feature extractors. Higher accuracy from co-learning with ImageNet feature extractors demonstrates the presence of relevant target information in these feature extractors and the ability of co-learning to distil such information. \label{tab: comparison_resnet}}
\end{table}

\begin{figure}
  \centering
  \begin{subfigure}{0.48\linewidth}
    \includegraphics[width=\linewidth]{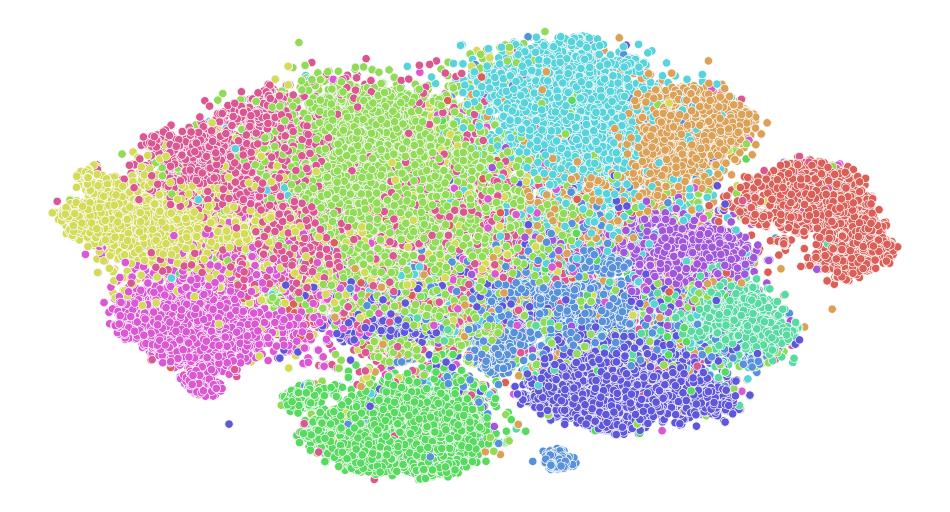}
    \caption{ImageNet ResNet-101}
  \end{subfigure}
  \hfill  
  \begin{subfigure}{0.48\linewidth}
    \includegraphics[width=\linewidth]{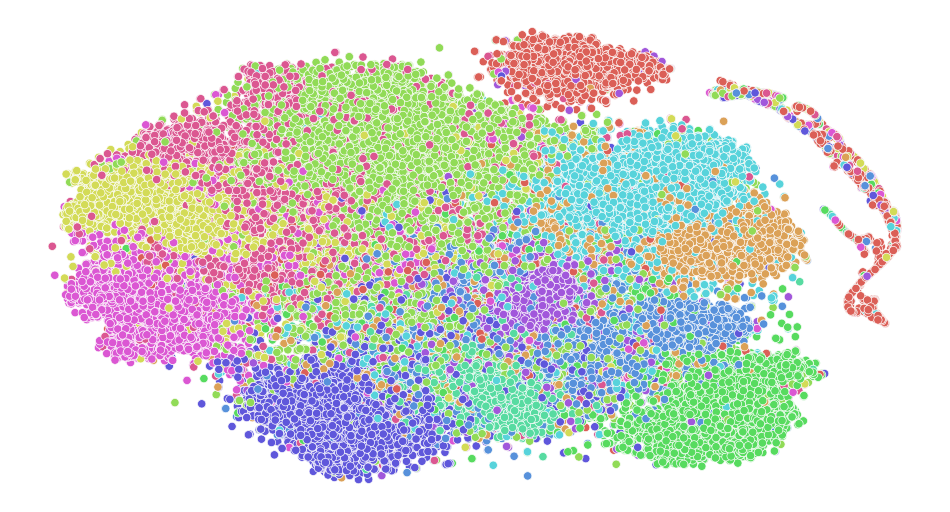}
    \caption{Source ResNet-101}
  \end{subfigure}

  \begin{subfigure}{0.48\linewidth}
    \includegraphics[width=\linewidth]{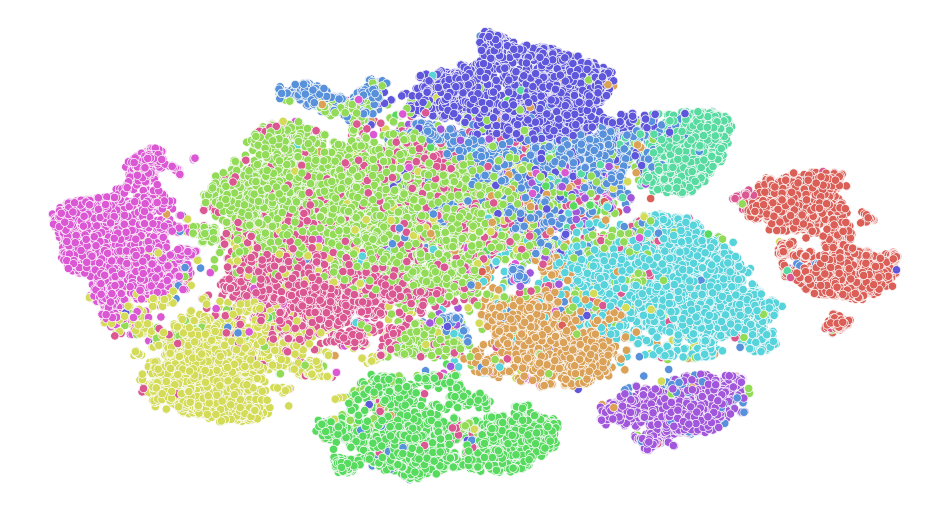}
    \caption{ImageNet Swin-B}
  \end{subfigure}
  \hfill    
  \begin{subfigure}{0.48\linewidth}
    \includegraphics[width=\linewidth]{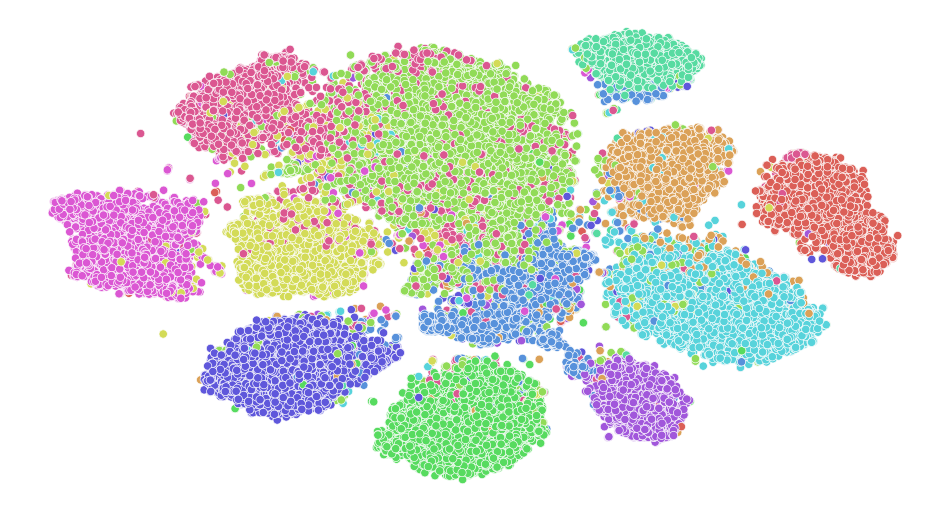}
    \caption{Adapted ResNet-101}
  \end{subfigure}

  \begin{subfigure}{0.48\linewidth}
    \includegraphics[width=\linewidth]{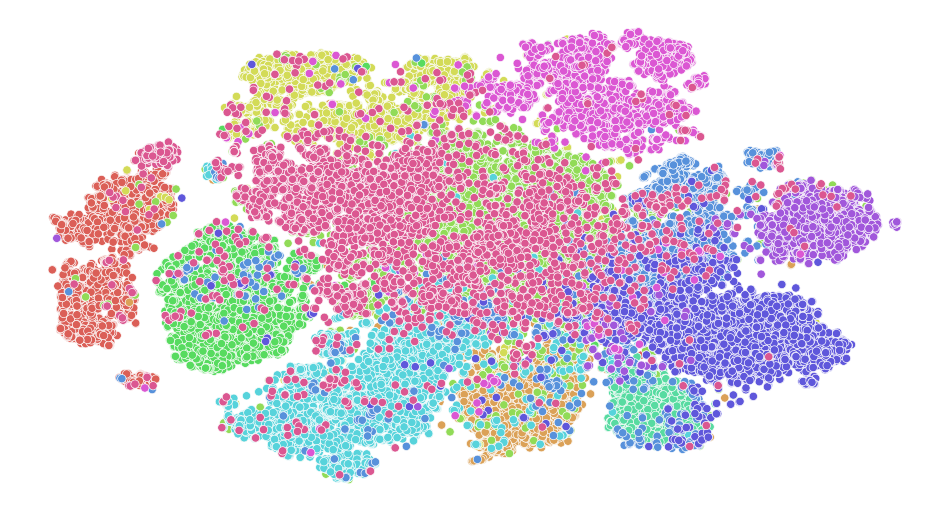}
    \caption{WIT CLIP}
  \end{subfigure}
  \hfill    
  \begin{subfigure}{0.48\linewidth}
    \includegraphics[width=\linewidth]{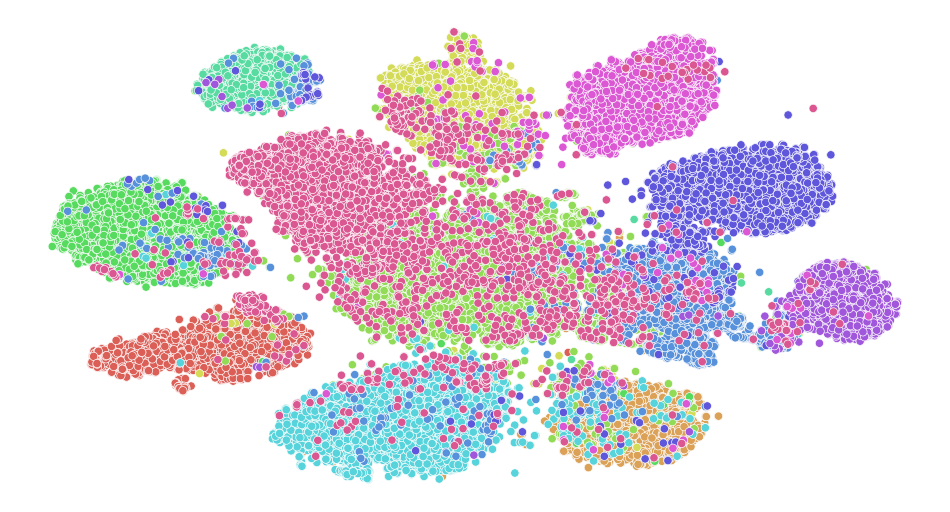}
    \caption{Adapted ResNet-101}
  \end{subfigure}

  \begin{subfigure}{0.8\linewidth}
    \includegraphics[width=\linewidth]{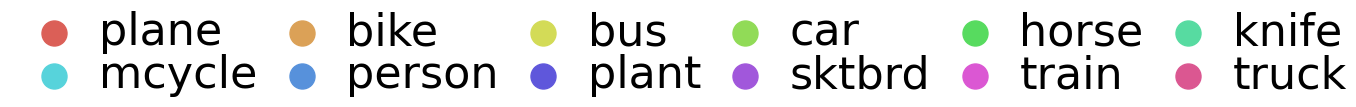}
  \end{subfigure}    
  
  \caption{t-SNE visualization of VisDA-C target domain features by (a) ImageNet-1k ResNet-101, (b) source-trained ResNet-101, (c) ImageNet-1k Swin-B, (d) source ResNet-101 adapted by co-learning with ImageNet-1k Swin-B, (e) WIT CLIP, and (f) source ResNet-101 adapted by co-learning with WIT CLIP. Features are extracted at the last pooling layer before the classification head. Samples are colored by class.}
  \label{fig: tsne}
\end{figure}

Secondly, the choice of pre-trained model can improve the robustness of input-to-feature mapping against covariate shift. In Table~\ref{tab: comparison_resnet} for the Office-Home dataset, ImageNet ResNet-101 has higher oracle target accuracy than ImageNet ResNet-50. Figure~\ref{fig: tsne} visually compares VisDA-C target domain features extracted at the last pooling layer before the classification head of several models. Swin~\citep{liu2021Swin} is a recent transformer architecture with strong representation learning ability. Even without training on VisDA-C images, the ImageNet Swin-B feature extractor produces more class-discriminative target representations than both ImageNet and source ResNet-101 feature extractors. The obervation is similar for CLIP pre-trained on the WebImage Text (WIT) dataset of image-text pairs. Furthermore, CLIP can generalize to different output label spaces just by assessing the text class labels.

\section{Proposed Strategy}
\label{sec: proposed strategy}

From our observations in Section~\ref{sec: role of pre-trained networks}, we propose to distil effective target domain information in pre-trained networks to generate improved pseudolabels to finetune the source model during target adaptation. We propose the `Co-learn' algorithm to integrate pre-trained vision encoders, and the `\colearnplus' algorithm to integrate the pre-trained vision-language CLIP model and its zero-shot classification decisions.

\subsection{Preliminaries}
\label{subsec: preliminaries}

We denote the source and target distributions as 
$\mathcal{P}_s$ and $\mathcal{P}_t$, and observations as $\mathcal{D}_s = \{(x_s^n, y_s^n)\}_{n=1}^{N_s}$ and $\mathcal{D}_t = \{(x_t^n, y_t^n)\}_{n=1}^{N_t}$, for image $x\in\mathcal{X}$ and one-hot classification label $y\in\mathcal{Y}$. The two domains have different data space $\mathcal{X}_s \neq \mathcal{X}_t$. For the default closed-set scenario, the two domains share the same label space $\mathcal{Y}_s = \mathcal{Y}_t$ with $|\mathcal{Y}|=L$ classes. For the more challenging open-set, partial-set and open-partial scenario, the label spaces do not match exactly. That is, $\mathcal{Y}_s \subset \mathcal{Y}_t$ for open-set, $\mathcal{Y}_t \subset \mathcal{Y}_s$ for partial-set, and $\mathcal{Y}_s \cap \mathcal{Y}_t \neq \emptyset$ and $\mathcal{Y}_s \cap \mathcal{Y}_t^c \neq \emptyset$ and $\mathcal{Y}_s^c \cap \mathcal{Y}_t \neq \emptyset$ for open-partial scenario. In SFDA setting, source data $\mathcal{D}_s$ and target labels $\{y_t^n\}_{n=1}^{N_t}$ are not accessible during adaptation. Knowledge on source data is captured by a source model.

The source model trained on $\mathcal{D}_s$ is composed of a feature extractor $f_s$ parameterized by $\Theta_s$ that yields learned representations $z_s(x)=f_s(x;\Theta_s)$, and a classifier $k_s$ parameterized by $\Psi_s$ that yields logits $g_s(x)=k_s(z_s(x);\Psi_s)$. Estimated class probabilities are obtained by $p_s(x)=\sigma(g_s(x))$ for softmax function $\sigma$ where $p_s(x)[i] = \frac{\exp(g_s(x)[i])}{\sum_{j=1}^L \exp{(g_s(x)[j])}}$ for class $i$, and the predicted class is $ \hat{y}_s = \arg\max_i p_s(x)[i]$. 

For a hypothesis $h\in\mathcal{H}$, we refer to $\epsilon_t(h, \ell_t) = E_{x\sim P_t} \epsilon(h(x), \ell_t(x))$ as the target risk of $h$ with respect to the true target labeling function $\ell_t$. To understand the relationship of $h$ and $l_t$ with the source model, we assume an error function $\epsilon$ such as $\epsilon(v,v')=|v-v'|^\alpha$ that satisfies triangle equality following \citep{BenDavid2010domainadaptation, blitzer2007bounds}, then
\begin{equation}
    \epsilon_t(h, \ell_t) \leq \epsilon_t(h, h_p) + \epsilon_t(h_p, \ell_t) \label{eqn: bound}
\end{equation}
where $h_p$ is a pseudolabeling function. The second term $\epsilon_t(h_p, \ell_t)$ is target pseudolabeling error by $h_p$, and the first term is minimized by $h=h_p$. Motivated by the bound in Equation~\ref{eqn: bound}, we propose an iterative strategy to improve the pseudolabeling function $h_p$ and to finetune $h$ towards $h_p$. Our pseudolabeling function leverages a pre-trained network, which is discarded after target adaptation.

\subsection{Two-branch framework}
\label{subsec: two-branch framework}

We propose a co-learning strategy to progressively adapt the source model $\{f_s,k_s\}$ with the pre-trained network $\{f_*,k_*\}$. The framework consists of two branches: (1) adaptation model branch $\{f_a,k_a\}$ initialized by source model $\{f_s,k_s\}$, (2) pre-trained model branch initialized by $f_*$ and a newly-estimated task classifier $q_*$ or $q^{\plus}_*$ for Co-learn or \colearnplus, respectively. We introduce the setup of the framework in this section, and describe the target adaptation procedure in Section~\ref{subsec: proposed sfda}.

\subsubsection{Integrating pre-trained vision encoder}
\label{subsubsec: colearn}

We denote the algorithm for co-learning with pre-trained vision encoders $f_*$ as `Co-learn'. A new task-specific classification head $q_*$ needs to be estimated for $f_*$. The original classifier $k_*$ from the pre-trained network is no longer suitable since the label space of interest has changed.

Inspired by the nearest-centroid-classifier (NCC) in \citep{liang2020shot}, we construct $q_*$ as a weighted nearest-centroid-classifier where the centroid $\mu_i$ for class $i$ is the sum of $f_*$ features, weighted by estimated class probabilities of the adaptation model $p_a(x) = \sigma(k_a(f_a(x;\Theta_a);\Psi_a))$:
\begin{align}
    & \mu_i = \frac{\sum_{x} p_a(x)[i] f_*(x;\Theta_*) / \|f_*(x;\Theta_*)\|}{\sum_{x} p_a(x)[i]} \label{eqn: pretrained_centroid} \\
    & g_*(x)[i] = q_*(f_*(x;\Theta_*))[i] = \frac{f_*(x;\Theta_*) \cdot \mu_i}{\|f_*(x;\Theta_*)\| \|\mu_i\|} \label{eqn: pretrained_logit} \\ 
    & p_*(x) = \sigma(g_*(x) / T) \label{eqn: pretrained_prob}
\end{align}
The weighted NCC leverages the target domain class-discriminative cluster structures in $f_*$ features. In Equation~\ref{eqn: pretrained_logit}, the logits $g_*(x)$ are computed by cosine similarity between the features and centroids. In Equation~\ref{eqn: pretrained_prob}, predictions $p_*(x)$ are sharpened with temperature $T = 0.01$ since the logits $g_*(x)$ are bounded in $[-1,1]$.

\subsubsection{Integrating pre-trained vision-language CLIP model}
\label{subsubsec: colearnplus}

We denote the algorithm for co-learning with the pre-trained vision-language CLIP model \citep{Radford2021clip} as `\colearnplus'. Note that while Co-learn in Section~\ref{subsubsec: colearn} can integrate the vision encoder component of CLIP into the co-learning framework, it does not consider the text encoder component. By additionally considering the CLIP text encoder in \colearnplus, we are able to create and leverage a zero-shot classifier $\tilde{k}_*$ suitable for the target domain label space. That is, the task-specific classification head $q^{\plus}_*$ is estimated under the guidance of zero-shot classification decisions from $\tilde{k}_*$.

We follow \citep{lin2023multimodality} to construct a NCC zero-shot classifier for CLIP. Let $\tilde{f}_*$ parameterized by $\tilde{\Theta}_*$ denote the CLIP text encoder, then $\tilde{f}_*(e; \tilde{\Theta}_*)$ is the text embedding for input text $e$. For the text class label $e_i$ of class $i$, we use the collection of 180 text templates $\{\tilde{t}_j\}_{j=1}^{180}$ from \citep{lin2023multimodality} to create a corresponding collection of text variations of $e_i$ i.e. $\{\tilde{t}_j(e_i)\}_{j=1}^{180}$ to capture different possible textual descriptions of $e_i$. For instance, for $e_i=``bike"$, the collection of text variations would include \textit{``bike"}, \textit{``a photo of a bike"}, \textit{``a rendering of a bike"}, and \textit{``a demonstration of a person using bike"}. The zero-shot centroid $\tilde{\mu}_i$ for class $i$ is then the average of the text embeddings of $\{\tilde{t}_j(e_i)\}_{j=1}^{180}$:
\begin{align}
    & \tilde{\mu}_i = \frac{\sum_{j=1}^{180} \tilde{f}_*(\tilde{t}_j(e_i); \tilde{\Theta}_*) / \|\tilde{f}_*(\tilde{t}_j(e_i); \tilde{\Theta}_*)\|}{180} \label{eqn: text_pretrained_centroid} \\
    & \tilde{g}_*(x)[i] = \tilde{k}_*(f_*(x;\Theta_*))[i] = \frac{f_*(x;\Theta_*) \cdot \tilde{\mu}_i}{\|f_*(x;\Theta_*)\| \|\tilde{\mu}_i\|} \label{eqn: text_pretrained_logit} \\ 
    & \tilde{p}_*(x) = \sigma(\tilde{g}_*(x) / \tilde{T}) \label{eqn: text_pretrained_prob}
\end{align}
In Equation~\ref{eqn: text_pretrained_logit}, the zero-shot logits $\tilde{g}_*(x)$ are computed
by cosine similarity between the CLIP image embeddings and zero-shot centroids obtained from CLIP text embeddings. 

We modify the formulation of the weighted NCC in Equation~\ref{eqn: pretrained_centroid}-\ref{eqn: pretrained_prob} such that CLIP zero-shot classifier outputs can guide the final prediction:
\begin{align}
    & \mu^{\plus}_i = \frac{\sum_{x} w^{\plus}(x)[i] f_*(x;\Theta_*) / \|f_*(x;\Theta_*)\|}{\sum_{x} w^{\plus}(x)[i]} \label{eqn: clip_pretrained_centroid} \\
    & \hspace{0.5cm} \text{where}\ w^{\plus}(x) = p_a(x) \odot \tilde{p}_*(x)  \nonumber \\
    & g^{\plus}_*(x)[i] = q_*^{\plus}(f_*(x;\Theta_*))[i] \nonumber \\
    & \hspace{0.5cm} = \alpha \frac{f_*(x;\Theta_*) \cdot \mu^{\plus}_i}{\|f_*(x;\Theta_*)\| \|\mu^{\plus}_i\|} + (1-\alpha) \tilde{g}_*(x)/\tilde{T}  \label{eqn: clip_pretrained_logit} \\ 
    & p^{\plus}_*(x) = \sigma(g^{\plus}_*(x) / T) \label{eqn: clip_pretrained_prob}
\end{align}
where $\odot$ denotes the element-wise product. CLIP zero-shot classifier outputs are applied as NCC weights in Equation~\ref{eqn: clip_pretrained_centroid}, and more directly at logit computation in Equation~\ref{eqn: clip_pretrained_logit}. The optimal strength of zero-shot guidance depends on the accuracy of the zero-shot classifier, which differs across datasets. In our experiments, we define two types of zero-shot guidance. For weak guidance, we set $\alpha=1$, and temperature $\tilde{T}=0.05$ according to zero-shot CLIP in \citep{deng2023uniood}. For strong guidance, we set $\alpha=0.5$ as in an ensemble classifier, and $\tilde{T}$ as the ratio of standard deviation of $\tilde{g}_*(x)$ to that of $\frac{f_*(x;\Theta_*) \cdot \mu^{\plus}_i}{\|f_*(x;\Theta_*)\| \|\mu^{\plus}_i\|}$ to balance the constribution of each classifier in the ensemble.

\subsection{Co-learning for adaptation}
\label{subsec: proposed sfda}

\begin{table}[tb]

\centering
\begin{adjustbox}{max width=0.95\columnwidth}
\begin{tabular}{*{4}{c}}
\toprule[1pt]\midrule[0.3pt]
$\mathbf{\hat{y}_a = \hat{y}_*}$    & $\mathbf{Conf(\hat{y}_a)} > \gamma$    & $\mathbf{Conf(\hat{y}_*)}>\gamma$  & \textbf{Pseudolabel} $\mathbf{\bar{y}}$ \\ \midrule
\checkmark  & \checkmark / \xmark   & \checkmark / \xmark   & $\hat{y}_a$ \\
\xmark      & \checkmark            & \checkmark            & - \\
\xmark      & \checkmark            & \xmark                & $\hat{y}_a$ \\
\xmark      & \xmark                & \checkmark            & $\hat{y}_*$ \\
\xmark      & \xmark                & \xmark                & - \\
\midrule[0.3pt]\bottomrule[1pt]
\end{tabular}
\end{adjustbox}
\caption{MatchOrConf pseudolabeling scheme with adaptation and pre-trained model predictions, $\hat{y}_a$ and $\hat{y}_*$, and confidence threshold $\gamma$. Dash (-) means no pseudolabel. \label{tab: pseudolabel_strategy}}
\end{table}

\begin{algorithm}[htb]
\caption{Co-learn for integrating pre-trained vision encoder\label{alg colearn}}

\textbf{Input:} Target images $\{x_t^n\}_{n=1}^{N_t}$;
source model $k_s \circ f_s$ parameterized by $(\Theta_s, \Psi_s)$; pre-trained vision encoder $f_*$ parameterized by $\Theta_*$; confidence threshold $\gamma$; learning rate $\eta$; \# episodes $I$;
\begin{algorithmic}[1]

\Procedure{Co-learn}{}
\State Initialize weights of adaption model branch $k_a \circ f_a$ by $(\Theta_a, \Psi_a) \leftarrow (\Theta_s, \Psi_s)$
\State Initialize pre-trained model branch classifier $q_*$ by computing centroids according to Equation~\ref{eqn: pretrained_centroid}

\For{episode $i=1:I$}
\State Construct pseudolabel dataset $\bar{\mathcal{D}}_t$ by MatchOrConf scheme with confidence threshold $\gamma$ in Table~\ref{tab: pseudolabel_strategy}
\State Compute objective $L_{co-learning}(\Theta_a)$ in Equation~\ref{eqn: colearn}
\State Update adaptation model branch weights $\Theta_a \leftarrow \Theta_a - \eta \nabla L_{co-learning}(\Theta_a)$
\State Update pre-trained model branch centroids in $q_*$ according to Equation~\ref{eqn: pretrained_centroid}
\EndFor
\EndProcedure

\State \Return Adaptation model $k_a \circ f_a$ with weights $(\Theta_a, \Psi_a)$
\end{algorithmic}
\end{algorithm}

\noindent\textbf{Adaptation procedure:}
We iterate between updates on the adaptation and pre-trained model branch, with each branch adapting to the target domain and producing more accurate predictions to improve the other branch. Each iteration of update on both branches is denoted as a co-learning episode. For the adaptation model branch, we freeze classifier $k_a$ and update feature extractor $f_a$ to rectify its initial bias towards the source domain. We finetune $f_a$ with cross-entropy loss on refined pseudolabeled samples, as elaborated in the subsequent paragraph. For the pre-trained model branch, we freeze feature extractor $f_*$ to preserve the target-compatible features and update classifier $q_*$ (or $q^{\plus}_*$ for \colearnplus). Since $f_*$ is not modified throughout the adaptation process, only a single pass of target data through $f_*$ is needed to construct a feature bank; previous works similarly required memory banks of source model features~\citep{Yang2022AttractingAD,yang2021gsfda,Yang2021ExploitingTI}. The centroids are updated by Equation~\ref{eqn: pretrained_centroid} (or Equation~\ref{eqn: clip_pretrained_centroid} for \colearnplus) using estimated class probabilities $p_a(x) = \sigma(k_a(f_a(x;\Theta_a);\Psi_a))$ from the current adaptation model $\{f_a, k_a\}$. Improved predictions from the adaptation model in each episode contribute to more accurate class centroids within the pre-trained model branch. 

\noindent\textbf{Refined pseudolabels:}
We improve pseudolabels each episode by integrating predictions from the two branches. For Co-learn and \colearnplus\ with weak zero-shot guidance, we apply the MatchOrConf scheme outlined in Table~\ref{tab: pseudolabel_strategy}. Denoting $\hat{y}_a$ and $\hat{y}_*$ as the adaptation and pre-trained model predictions, respectively, the refined pseudolabel $\bar{y}$ assigned is $\hat{y}_a (=\hat{y}_*)$ if the predicted classes match, and the predicted class with higher confidence level otherwise. The confidence level is determined by a threshold $\gamma$. For \colearnplus\ with strong zero-shot guidance, we assign refined pseudolabels $\bar{y}$ according to pretrained model predictions $\hat{y}_*$ with confidence above $\gamma$. Remaining samples are not pseudolabeled. The co-learning objective for the adaptation model is
\begin{equation}
    L_{co-learning}(\Theta_a) = -\sum_{(x,\bar{y})\in \bar{D}_t} \bar{y} \cdot \log( p_a(x) ) \label{eqn: colearn}
\end{equation}
where $p_a(x) = \sigma(k_a( f_a(x; \Theta_a) ; \Psi_a))$ and $\bar{\mathcal{D}}_t$ is the pseudolabeled target dataset. Algorithm~\ref{alg colearn} and \ref{alg colearnplus} summarize Co-learn and \colearnplus\ respectively. We can incorporate the co-learning strategy into existing SFDA methods by replacing the pseudolabels used with our co-learned pseudolabels or by adding $L_{co-learning}$ to the learning objective.

\begin{algorithm}[tb]
\caption{\colearnplus\ for integrating pre-trained vision-language CLIP model \label{alg colearnplus}}

\textbf{Input:} Target images $\{x_t^n\}_{n=1}^{N_t}$;
source model $k_s \circ f_s$ parameterized by $(\Theta_s, \Psi_s)$; pre-trained CLIP vision encoder $f_*$ parameterized by $\Theta_*$; pre-trained CLIP text encoder $\tilde{f}_*$ parameterized by $\tilde{\Theta}_*$, confidence threshold $\gamma$; learning rate $\eta$; \# episodes $I$;
\begin{algorithmic}[1]

\Procedure{\colearnplus}{}
\State Initialize weights of adaption model branch $k_a \circ f_a$ by $(\Theta_a, \Psi_a) \leftarrow (\Theta_s, \Psi_s)$
\State Initialize pre-trained model branch classifier $q^{\plus}_*$ by computing centroids according to Equation~\ref{eqn: clip_pretrained_centroid}

\For{episode $i=1:I$}
\State Construct pseudolabel dataset $\bar{\mathcal{D}}_t$ by MatchOrConf scheme with confidence threshold $\gamma$ in Table~\ref{tab: pseudolabel_strategy} for weak zero-shot guidance, or pre-trained model branch predictions with confidence threshold $\gamma$ for strong zero-shot guidance
\State Compute objective $L_{co-learning}(\Theta_a)$ in Equation~\ref{eqn: colearn}
\State Update adaptation model branch weights $\Theta_a \leftarrow \Theta_a - \eta \nabla L_{co-learning}(\Theta_a)$
\State Update pre-trained model branch centroids in $q^{\plus}_*$ according to Equation~\ref{eqn: clip_pretrained_centroid}
\EndFor
\EndProcedure

\State \Return Adaptation model $k_a \circ f_a$ with weights $(\Theta_a, \Psi_a)$
\end{algorithmic}
\end{algorithm}

After target adaptation, the pre-trained model branch is discarded, only the adaptation model branch is retained. Therefore, no additional computation needs to be incurred during inference.

\section{Experiments and Results}
\label{sec: experiments and results}

We evaluate on 4 benchmark image classification datasets for domain adaptation. We describe experimental setups in Section~\ref{subsec: experimental setups} and results in Section~\ref{subsec: results}.

\subsection{Experimental setups}
\label{subsec: experimental setups}

\noindent\textbf{Datasets.}
\textbf{Office-31}~\citep{Saenko2010AdaptingVC} has 31 categories of office objects in 3 domains: Amazon (A), Webcam (W) and DSLR (D).
\textbf{Office-Home}~\citep{Venkateswara2017DeepHN} has 65 categories of everyday objects in 4 domains: Art (A), Clipart (C), Product (P) and Real World (R).
\textbf{VisDA-C}~\citep{Peng2017VisDATV} is a popular 12-class dataset for evaluating synthetic-to-real shift, with synthetic rendering of 3D models in source domain and Microsoft COCO real images in target domain.
\textbf{DomainNet}~\citep{peng2019moment} is a challenging dataset with a total of 6 domains and 345 classes. Following \cite{litrico2023uncertainty}, we evaluate on 4 domains: Clipart (C), Painting (P), Real (R) and Sketch (S) with a subset of 126 classes. 
We report classification accuracy on the target domain for all domain pairs in Office-31 and Office-Home, and average per-class accuracy on the real domain in VisDA-C.

\begin{table*}[tb]
\centering
\begin{adjustbox}{max width=0.80\textwidth}
\begin{tabular}{l*{7}{c}l}
\toprule[1pt]\midrule[0.3pt]
\textbf{Method}                 & \textbf{SF} & $\mathbf{A\rightarrow D}$ & $\mathbf{A\rightarrow W}$ & $\mathbf{D\rightarrow A}$  
                                & $\mathbf{D\rightarrow W}$ & $\mathbf{W\rightarrow A}$ & $\mathbf{W\rightarrow D}$ &\textbf{Avg} \\ \midrule                                
MDD \citep{zhang2019mdd}         & \xmark & 93.5  & 94.5  & 74.6  & 98.4  & 72.2  & \underline{\textbf{100.0}} & 88.9 \\
GVB-GD \citep{cui2020gvb}        & \xmark & 95.0  & 94.8  & 73.4  & 98.7  & 73.7  & \underline{\textbf{100.0}} & 89.3 \\
MCC \citep{jin2020mcc}           & \xmark & 95.6  & 95.4  & 72.6  & 98.6  & 73.9  & \underline{\textbf{100.0}} & 89.4 \\
GSDA \citep{hu2020gsda}          & \xmark & 94.8  & \textbf{95.7}  & 73.5  & 99.1  & 74.9  & \underline{\textbf{100.0}} & 89.7 \\
CAN \citep{kang2019can}          & \xmark & 95.0  & 94.5  & \textbf{78.0}  & 99.1  & 77.0  & 99.8  & 90.6 \\
SRDC \citep{tang2020srdc}        & \xmark & \textbf{95.8}  & \textbf{95.7}  & 76.7  & \textbf{99.2}  & \textbf{77.1}  & \underline{\textbf{100.0}} & \textbf{90.8} \\ \midrule
Source Only                     & \cmark & 81.9  & 78.0  & 59.4  & 93.6  & 63.4  & 98.8  & 79.2 \\
Co-learn (w/ ResNet-50)         & \cmark & 93.6  & 90.2  & 75.7  & 98.2  & 72.5  & 99.4  & 88.3 \\
Co-learn (w/ Swin-B)            & \cmark & \textbf{97.4}  & \textbf{98.2}  & \textbf{84.5}  & \textbf{99.1}  & \textbf{82.2}  & \underline{\textbf{100.0}} & \underline{\textbf{93.6}} \\ \hdashline
CLIP Zero-shot            & \cmark & 87.8 & 89.2 & 85.5 & 89.2 & \underline{\textbf{85.5}} & 87.8 & 87.5\\
Co-learn (w/ CLIP) & \cmark & 99.2 & \underline{\textbf{99.7}} & 85.3 & 99.1 & 83.2 & \underline{\textbf{100.0}} & 94.4 \\
\colearnplus (w/ CLIP) & \cmark & \underline{\textbf{99.6}} & 99.0 & \underline{\textbf{86.3}} & 99.1 & 84.8 & \underline{\textbf{100.0}} & \underline{\textbf{94.8}}\\ \midrule
SHOT$^\dagger$ \citep{liang2020shot}         & \cmark & 95.0 & 90.4 & 75.2 & \textbf{98.9} & 72.8 & 99.8 & 88.7 \\
\quad w/ Co-learn (w/ ResNet-50)            &        & 94.2 & 90.2 & 75.7 & 98.2 & 74.4 & \underline{\textbf{100.0}} & 88.8 $\green{(\uparrow 0.1)}$ \\
\quad w/ Co-learn (w/ Swin-B)               &        & \textbf{95.8} & \textbf{95.6} & \textbf{78.5} & \textbf{98.9} & \textbf{76.7} & 99.8 & \textbf{90.9} $\green{(\uparrow 2.2)}$ \\ \hdashline
\quad w/ Co-learn (w/ CLIP)                 &        & 95.6 & 92.8 & 77.4 & \textbf{98.7} & 77.5 & \underline{\textbf{100.0}} & 90.4 $\green{(\uparrow 1.7)}$\\
\quad w/ \colearnplus (w/ CLIP)             &        & \textbf{96.2} & \textbf{94.7} & \textbf{78.3} & 98.2 & \textbf{77.6} & 99.8 & \textbf{90.8} $\green{(\uparrow 2.1)}$\\ \midrule
SHOT++$^\dagger$ \citep{liang2021shotplus}   & \cmark & 95.6 & 90.8 & 76.0 & 98.2 & 74.6 & \underline{\textbf{100.0}} & 89.2 \\
\quad w/ Co-learn (w/ ResNet-50)            &        & 95.0 & 90.7 & 76.4 & 97.7 & 74.9 & 99.8 & 89.1 $\red{(\downarrow 0.1)}$ \\
\quad w/ Co-learn (w/ Swin-B)               &        & \textbf{96.6} & \textbf{93.8} & \textbf{79.8} & \textbf{98.9} & \textbf{78.0} & \underline{\textbf{100.0}} & \textbf{91.2} $\green{(\uparrow 2.0)}$ \\ \hdashline
\quad w/ Co-learn (w/ CLIP)                 &        & 95.4 & \textbf{95.2} & \textbf{78.9} & \textbf{98.9} & \textbf{78.5} & \underline{\textbf{100.0}} & \textbf{91.1} $\green{(\uparrow 1.9)}$ \\
\quad w/ \colearnplus (w/ CLIP)             &        & \textbf{96.8} & 92.3 & 78.3 & 98.4 & 77.8 & 99.8 & 90.6 $\green{(\uparrow 1.4)}$\\ \midrule
NRC$^\dagger$ \citep{Yang2021ExploitingTI}   & \cmark & 92.0 & 91.6 & 74.5 & 97.9 & 74.8 & \underline{\textbf{100.0}} & 88.5 \\
\quad w/ Co-learn (w/ ResNet-50)            &        & 95.4 & 89.9 & 76.5 & 98.0 & 76.0 & 99.8 & 89.3 $\green{(\uparrow 0.8)}$\\
\quad w/ Co-learn (w/ Swin-B)               &        & \textbf{96.2} & \textbf{94.3} & \textbf{79.1} & \textbf{98.7} & \textbf{78.5} & \underline{\textbf{100.0}} & \textbf{91.1} $\green{(\uparrow 2.6)}$\\ \hdashline
\quad w/ Co-learn (w/ CLIP)                 &         & \textbf{96.6} & \textbf{96.3} & 78.0 & 98.5 & 77.9 & \underline{\textbf{100.0}} & 91.2 $\green{(\uparrow 2.7)}$\\
\quad w/ \colearnplus (w/ CLIP)             &         & 96.4 & 95.8 & \textbf{78.7} & \textbf{98.9} & \textbf{78.5} & 99.8 & \textbf{91.4} $\green{(\uparrow 2.9)}$\\ \midrule
AaD$^\dagger$ \citep{Yang2022AttractingAD}   & \cmark & 94.4 & 93.3 & 75.9 & 98.4 & 76.3 & 99.8 & 89.7 \\
\quad w/ Co-learn (w/ ResNet-50)            &        & 96.6 & 92.5 & 77.3 & 98.9 & 76.6 & 99.8 & 90.3 $\green{(\uparrow 0.6)}$\\
\quad w/ Co-learn (w/ Swin-B)               &        & \textbf{97.6} & \textbf{98.7} & \textbf{82.1} & \underline{\textbf{99.3}} & \textbf{80.1} & \underline{\textbf{100.0}} & \textbf{93.0} $\green{(\uparrow 3.3)}$\\ \hdashline
\quad w/ Co-learn (w/ CLIP)                 &        & 95.2 & 96.2 & 78.5 & 98.6 & 79.7 & \underline{\textbf{100.0}} & 91.4 $\green{(\uparrow 1.7)}$\\
\quad w/ \colearnplus (w/ CLIP)             &        & \textbf{98.0} & \textbf{97.7} & \textbf{81.3} & \textbf{99.1} & \textbf{82.1} & \underline{\textbf{100.0}} & \textbf{93.0} $\green{(\uparrow 3.3)}$\\
\midrule[0.3pt]\bottomrule[1pt]
\end{tabular}
\end{adjustbox}
\caption{Office-31: 31-class classification accuracy of adapted ResNet-50. The source model is initialized with ImageNet-1k ResNet-50 weights. For proposed strategy, the pre-trained network used for co-learning is given in parenthesis: CLIP is pre-trained on WIT, and the rest are pre-trained on ImageNet-1k (i.e. no new data is introduced). CLIP Zero-shot utilizes the text template-based classifier from \citep{lin2023multimodality}.} SF denotes source-free. $\dagger$ denotes reproduced results. \label{tab: office31_results}
\end{table*}

\begin{table*}[tb]
\centering
 \setlength{\tabcolsep}{2pt}
\begin{adjustbox}{max width=\textwidth}
\begin{tabular}{l*{13}{c}l}
\toprule[1pt]\midrule[0.3pt]
\textbf{Method}                 & \textbf{SF} & $\mathbf{A\rightarrow C}$ & $\mathbf{A\rightarrow P}$ & $\mathbf{A\rightarrow R}$  
                                & $\mathbf{C\rightarrow A}$ & $\mathbf{C\rightarrow P}$ & $\mathbf{C\rightarrow R}$ 
                                & $\mathbf{P\rightarrow A}$ & $\mathbf{P\rightarrow C}$ & $\mathbf{P\rightarrow R}$
                                & $\mathbf{R\rightarrow A}$ & $\mathbf{R\rightarrow C}$ & $\mathbf{R\rightarrow P}$ & \textbf{Avg}\\ \midrule
GSDA \citep{hu2020gsda}          & \xmark & \textbf{61.3}  & 76.1  & 79.4  & 65.4  & 73.3  & 74.3  & 65.0  & 53.2  & 80.0  & 72.2  & 60.6  & 83.1  & 70.3 \\
GVB-GD \citep{cui2020gvb}        & \xmark & 57.0  & 74.7  & 79.8  & 64.6  & 74.1  & 74.6  & 65.2  & 55.1  & 81.0  & 74.6  & 59.7  & 84.3  & 70.4 \\
RSDA \citep{gu2020rsda}          & \xmark & 53.2  & \textbf{77.7}  & \textbf{81.3}  & 66.4  & 74.0  & 76.5  & 67.9  & 53.0  & \textbf{82.0}  & 75.8  & 57.8  & 85.4  & 70.9 \\
TSA \citep{Li2021TSA}            & \xmark & 57.6  & 75.8  & 80.7  & 64.3  & 76.3  & 75.1  & 66.7  & 55.7  & 81.2  & 75.7  & 61.9  & 83.8  & 71.2 \\
SRDC \citep{tang2020srdc}        & \xmark & 52.3  & 76.3  & 81.0  & \textbf{69.5}  & 76.2  & 78.0  & \textbf{68.7}  & 53.8  & 81.7  & 76.3  & 57.1  & 85.0  & 71.3 \\
FixBi \citep{na2021fixbi}        & \xmark & 58.1  & 77.3  & 80.4  & 67.7  & \textbf{79.5}  & \textbf{78.1}  & 65.8  & \textbf{57.9}  & 81.7  & \textbf{76.4}  & \textbf{62.9}  & \textbf{86.7}  & \textbf{72.7} \\ \midrule
Source Only                     & \cmark & 43.5  & 67.1  & 74.2  & 51.5  & 62.2  & 63.3  & 51.4  & 40.7  & 73.2  & 64.6  & 45.8  & 77.6  & 59.6 \\
Co-learn (w/ ResNet-50)         & \cmark & 51.8  & 78.9  & 81.3  & 66.7  & 78.8  & 79.4  & 66.3  & 50.0  & 80.6  & 71.1  & 53.7  & 81.3  & 70.0\\
Co-learn (w/ Swin-B)            & \cmark & \textbf{69.6}  & \textbf{89.5}  & \textbf{91.2}  & \textbf{82.7}  & \textbf{88.4}  & \underline{\textbf{91.3}}  & \textbf{82.6}  & \textbf{68.5}  & \textbf{91.5}  & \textbf{82.8}  & \textbf{71.3}  & \textbf{92.1}  & \textbf{83.5} \\ \hdashline
CLIP Zero-shot            & \cmark & 72.6 & 86.5 & 85.2 & 81.8 & 86.5 & 85.2 & 81.8 & 72.6 & 85.2 & 81.8 & 72.6 & 86.5 & 81.5\\
Co-learn (w/ CLIP)              & \cmark & 77.2 & 90.4 & 91.0 & 77.1 & 88.1 & 90.0 & 76.6 & 72.5 & 90.1 & 82.0 & 79.6 & 93.0 & 84.0 \\
\colearnplus (w/ CLIP)          & \cmark & \underline{\textbf{80.0}} & \underline{\textbf{91.2}} & \underline{\textbf{91.8}} & \underline{\textbf{83.4}} & \underline{\textbf{92.7}} & \underline{\textbf{91.3}} & \underline{\textbf{83.4}} & \underline{\textbf{78.9}} & \underline{\textbf{92.0}} & \underline{\textbf{85.5}} & \underline{\textbf{80.6}} & \underline{\textbf{94.7}} & \underline{\textbf{87.1}}\\ \midrule
SHOT$^\dagger$ \citep{liang2020shot}         & \cmark & 55.8 & 79.6 & 82.0 & 67.4 & 77.9 & 77.9 & 67.6 & 55.6 & 81.9 & 73.3 & 59.5 & 84.0 & 71.9\\
\quad w/ Co-learn (w/ ResNet-50)            &        & 56.3 & 79.9 & 82.9 & 68.5 & 79.6 & 78.7 & 68.1 & 54.8 & 82.5 & 74.5 & 59.0 & 83.6 & 72.4 $\green{(\uparrow 0.5)}$ \\
\quad w/ Co-learn (w/ Swin-B)               &        & \textbf{61.7} & \textbf{82.9} & \textbf{85.3} & \textbf{72.7} & \textbf{80.5} & \textbf{82.0} & \textbf{71.6} & \textbf{60.4} & \textbf{84.5} & \textbf{76.0} & \textbf{64.3} & \textbf{86.7} & \textbf{75.7} $\green{(\uparrow 3.8)}$ \\ \hdashline
\quad w/ Co-learn (w/ CLIP)                 &        & \textbf{63.1} & \textbf{83.1} & 84.6 & \textbf{72.0} & \textbf{81.9} & \textbf{82.0} & 70.8 & 60.4 & 83.8 & \textbf{75.9} & \textbf{66.1} & 86.1 & \textbf{75.8} $\green{(\uparrow 3.9)}$\\
\quad w/ \colearnplus (w/ CLIP)             &         & 62.2 & \textbf{83.1} & \textbf{84.9} & 71.5 & 81.7 & 81.7 & \textbf{70.9} & \textbf{61.9} & \textbf{84.1} & \textbf{75.9} & 65.5 & \textbf{86.6} & \textbf{75.8} $\green{(\uparrow 3.9)}$ \\ \midrule
SHOT++$^\dagger$ \citep{liang2021shotplus}   & \cmark & 57.1 & 79.5 & 82.6 & 68.5 & 79.5 & 78.6 & 68.3 & 56.1 & 82.9 & 74.0 & 59.8 & 85.0 & 72.7 \\
\quad w/ Co-learn (w/ ResNet-50)            &        & 57.7 & 81.1 & 84.0 & 69.2 & 79.8 & 79.2 & 69.1 & 57.7 & 82.9 & 73.7 & 60.1 & 85.0 & 73.3 $\green{(\uparrow 0.6)}$ \\
\quad w/ Co-learn (w/ Swin-B)               &        & \textbf{63.7} & \textbf{83.0} & \textbf{85.7} & \textbf{72.6} & \textbf{81.5} & \textbf{83.8} & \textbf{72.0} & \textbf{59.9} & \textbf{85.3} & \textbf{76.3} & \textbf{65.3} & \textbf{86.6} & \textbf{76.3} $\green{(\uparrow 3.6)}$ \\ \hdashline
\quad w/ Co-learn (w/ CLIP)                 &        & \textbf{63.6} & \textbf{83.6} & \textbf{84.8} & 71.4 & \textbf{81.7} & 81.7 & 70.2 & 58.7 & \textbf{84.4} & 76.3 & \textbf{66.3} & 86.3 & 75.8 $\green{(\uparrow 3.1)}$\\
\quad w/ \colearnplus (w/ CLIP)             &        & 62.5 & 83.5 & 84.5 & \textbf{72.7} & 81.5 & \textbf{83.2} & \textbf{71.1} & \textbf{61.3} & 84.2 & \textbf{76.6} & 65.9 & \textbf{86.4} & \textbf{76.1} $\green{(\uparrow 3.4)}$ \\ \midrule
NRC$^\dagger$ \citep{Yang2021ExploitingTI}   & \cmark & 58.0 & 79.3 & 81.8 & 70.1 & 78.7 & 78.7 & 63.5 & 57.0 & 82.8 & 71.6 & 58.2 & 84.3 & 72.0 \\
\quad w/ Co-learn (w/ ResNet-50)            &        & 56.1 & 80.3 & 83.0 & 70.3 & 81.3 & 80.9 & 67.7 & 53.9 & 83.7 & 72.5 & 57.9 & 83.4 & 72.6 $\green{(\uparrow 0.6)}$ \\
\quad w/ Co-learn (w/ Swin-B)               &        & \textbf{67.8} & \textbf{86.4} & \textbf{89.1} & \textbf{80.7} & \textbf{87.5} & \textbf{89.3} & \textbf{77.8} & \textbf{68.8} & \textbf{89.7} & \textbf{81.6} & \textbf{68.7} & \textbf{89.9} & \textbf{81.4} $\green{(\uparrow 9.4)}$ \\ \hdashline
\quad w/ Co-learn (w/ CLIP)                 &         & 72.2 & 87.6 & 88.4 & 77.8 & 87.3 & \textbf{88.3} & 77.9 & 70.7 & 89.4 & 79.9 & 74.2 & 90.7 & 82.0 $\green{(\uparrow 10.0)}$\\
\quad w/ \colearnplus (w/ CLIP)             &         & \textbf{76.4} & \textbf{88.8} & \textbf{88.6} & \textbf{82.4} & \textbf{89.1} & 88.2 & \textbf{81.0} & \textbf{75.2} & \textbf{89.7} & \textbf{82.4} & \textbf{76.3} & 91.9 & \textbf{84.2} $\green{(\uparrow 12.2)}$\\ \midrule
AaD$^\dagger$ \citep{Yang2022AttractingAD}   & \cmark & 58.7 & 79.8 & 81.4 & 67.5 & 79.4 & 78.7 & 64.7 & 56.8 & 82.5 & 70.3 & 58.0 & 83.3 & 71.8 \\
\quad w/ Co-learn (w/ ResNet-50)            &        & 57.7 & 80.4 & 83.3 & 70.1 & 80.1 & 80.6 & 66.6 & 55.5 & 84.1 & 72.1 & 57.6 & 84.3 & 72.7 $\green{(\uparrow 0.9)}$ \\
\quad w/ Co-learn (w/ Swin-B)               &        & \textbf{65.1} & \textbf{86.0} & \textbf{87.0} & \textbf{76.8} & \textbf{86.3} & \textbf{86.5} & \textbf{74.4} & \textbf{66.1} & \textbf{87.7} & \textbf{77.9} & \textbf{66.1} & \textbf{88.4} & \textbf{79.0} $\green{(\uparrow 7.2)}$\\ \hdashline
\quad w/ Co-learn (w/ CLIP)                 &         & 66.4 & 85.3 & \textbf{87.0} & 74.7 & 87.1 & 85.6 & 73.0 & 66.8 & 85.9 & 76.4 & 67.2 & 89.8 & 78.8 $\green{(\uparrow 7.0)}$\\
\quad w/ \colearnplus (w/ CLIP)             &         & \textbf{71.9} & \textbf{88.1} & 86.7 & \textbf{79.0} & \textbf{88.9} & \textbf{86.2} & \textbf{75.7} & \textbf{70.7} & \textbf{87.8} & \textbf{79.2} & \textbf{72.5} & \textbf{90.3} & \textbf{81.4} $\green{(\uparrow 9.6)}$\\
\midrule[0.3pt]\bottomrule[1pt]
\end{tabular}
\end{adjustbox}
\caption{Office-Home: 65-class classification accuracy of adapted ResNet-50. The source model is initialized with ImageNet-1k ResNet-50 weights. For proposed strategy, the pre-trained network used for co-learning is given in parenthesis: CLIP is pre-trained on WIT, and the rest are pre-trained on ImageNet-1k (i.e. no new data is introduced). CLIP Zero-shot utilizes the text template-based classifier from \citep{lin2023multimodality}.} SF denotes source-free. $\dagger$ denotes reproduced results. \label{tab: officehome_results}
\end{table*}

\begin{table*}[tb]
\centering
\setlength{\tabcolsep}{3pt}
\begin{adjustbox}{max width=\textwidth}
\begin{tabular}{l*{13}{c}l}
\toprule[1pt]\midrule[0.3pt]
\textbf{Method}                 & \textbf{SF} & \textbf{plane} & \textbf{bike} & \textbf{bus} & \textbf{car} & \textbf{horse} & \textbf{knife} & \textbf{mcycle} & \textbf{person} & \textbf{plant} & \textbf{sktbrd} & \textbf{train} & \textbf{truck} & \textbf{Avg}\\ \midrule
SFAN \citep{Xu2019sfan}          & \xmark & 93.6  & 61.3  & 84.1  & 70.6  & 94.1  & 79.0  & 91.8  & 79.6  & 89.9  & 55.6  & 89.0  & 24.4  & 76.1 \\
MCC \citep{jin2020mcc}           & \xmark & 88.7  & 80.3  & 80.5  & 71.5  & 90.1  & 93.2  & 85.0  & 71.6  & 89.4  & 73.8  & 85.0  & 36.9  & 78.8 \\
STAR \citep{lu2020star}          & \xmark & 95.0  & 84.0  & 84.6  & 73.0  & 91.6  & 91.8  & 85.9  & 78.4  & 94.4  & 84.7  & 87.0  & 42.2  & 82.7 \\
SE \citep{french2018se}          & \xmark & 95.9  & 87.4  & 85.2  & 58.6  & 96.2  & 95.7  & 90.6  & 80.0  & 94.8  & 90.8  & 88.4  & 47.9  & 84.3 \\
CAN \citep{kang2019can}          & \xmark & \textbf{97.0}  & 87.2  & 82.5  & 74.3  & \textbf{97.8}  & \textbf{96.2}  & 90.8  & 80.7  & 96.6  & \textbf{96.3}  & 87.5  & \textbf{59.9}  & \textbf{87.2} \\
FixBi \citep{na2021fixbi}        & \xmark & 96.1  & \textbf{87.8}  & \textbf{90.5}  & \underline{\textbf{90.3}}  & 96.8  & 95.3  & \textbf{92.8}  & \underline{\textbf{88.7}}  & \textbf{97.2}  & 94.2  & \textbf{90.9}  & 25.7  & \textbf{87.2} \\ \midrule
Source Only                     & \cmark & 51.5  & 15.3  & 43.4  & 75.4  & 71.2  & 6.8   & 85.5  & 18.8  & 49.4  & 46.4  & 82.1  & 5.4   & 45.9 \\
Co-learn (w/ ResNet-101)        & \cmark & 96.5  & 78.9  & 77.5  & 75.7  & 94.6  & 95.8  & 89.1  & 77.7  & 90.5  & 91.0  & 86.2  & \textbf{51.5}  & 83.7 \\
Co-learn (w/ Swin-B)            & \cmark & \textbf{99.0}  & \textbf{90.0}  & \textbf{84.2}  & \textbf{81.0}  & \textbf{98.1}  & \textbf{97.9}  & \textbf{94.9}  & \textbf{80.1}  & \textbf{94.8}  & \textbf{95.9}  & \textbf{94.4}  & 48.1  & \textbf{88.2} \\ \hdashline
CLIP Zero-shot            & \cmark & \underline{\textbf{99.6}} & 93.4 & \underline{\textbf{93.3}} & 73.5 & \underline{\textbf{99.7}} & 97.0 & \underline{\textbf{97.2}} & 73.0 & 88.6 & \underline{\textbf{99.2}} & \underline{\textbf{97.3}} & \underline{\textbf{70.2}} & 90.2 \\
Co-learn (w/ CLIP)              & \cmark & 98.9 & 93.2 & 81.0 & \textbf{83.0} & 98.6 & 98.8 & 95.7 & \textbf{84.8} & \textbf{94.8} & 97.3 & 95.1 & 41.6 & 88.6 \\
\colearnplus (w/ CLIP)          & \cmark & \underline{\textbf{99.6}} & \underline{\textbf{94.6}} & 90.9 & 77.8 & 99.6 & \underline{\textbf{99.0}} & 96.4 & 80.1 & 90.0 & \underline{\textbf{99.2}} & 96.3 & 70.1 & \underline{\textbf{91.1}} \\ \midrule
SHOT$^\dagger$ \citep{liang2020shot}         & \cmark & 95.3 & 87.1 & 79.1 & 55.1 & 93.2 & 95.5 & 79.5 & 79.6 & 91.6 & 89.5 & 87.9 & 56.0 & 82.4 \\
\quad w/ Co-learn (w/ ResNet-101)           &        & 94.9 & 84.8 & 77.7 & \textbf{63.0} & 94.1 & 95.6 & 85.6 & 81.0 & \textbf{93.0} & \textbf{92.2} & 86.4 & 60.4 & 84.1 $\green{(\uparrow 1.7)}$ \\
\quad w/ Co-learn (w/ Swin-B)               &        & \textbf{96.0} & \textbf{88.1} & \textbf{81.0} & \textbf{63.0} & \textbf{94.3} & \textbf{95.9} & \textbf{87.1} & \textbf{81.8} & 92.8 & 91.9 & \textbf{90.1} & \textbf{60.5} & \textbf{85.2} $\green{(\uparrow 2.8)}$ \\ \hdashline
\quad w/ Co-learn (w/ CLIP)                 &        & 96.3 & 89.8 & \textbf{83.8} & 63.0 & 95.6 & 96.7 & 88.4 & 82.1 & \textbf{91.7} & 91.4 & 88.6 & \textbf{62.2} & 85.8 $\green{(\uparrow 3.4)}$ \\
\quad w/ \colearnplus (w/ CLIP)             &        & \textbf{97.2} & \textbf{91.3} & \textbf{83.8} & \textbf{69.1} & \textbf{97.1} & \textbf{98.0} & \textbf{88.9} & \textbf{83.0} & 91.5 & \textbf{94.6} & \textbf{89.3} & 57.6 & \textbf{86.8} $\green{(\uparrow 4.4)}$\\ \midrule
SHOT++$^\dagger$ \citep{liang2021shotplus}   & \cmark & 94.5 & 88.5 & \textbf{90.4} & 84.6 & \textbf{97.9} & \textbf{98.6} & 91.9 & 81.8 & 96.7 & 91.5 & 93.8 & 31.3 & 86.8 \\
\quad w/ Co-learn (w/ ResNet-101)           &        & 97.9 & 88.6 & 86.8 & \textbf{86.7} & \textbf{97.9} & \textbf{98.6} & \textbf{92.4} & 83.6 & 97.4 & 92.5 & 94.4 & 32.5 & 87.4 $\green{(\uparrow 0.6)}$ \\
\quad w/ Co-learn (w/ Swin-B)               &        & \textbf{98.0} & \textbf{91.1} & 88.6 & 83.2 & 97.8 & 97.8 & 92.0 & \textbf{85.8} & \textbf{97.6} & \textbf{93.2} & \textbf{95.0} & \textbf{43.5} & \textbf{88.6} $\green{(\uparrow 1.8)}$ \\ \hdashline
\quad w/ Co-learn (w/ CLIP)                 &        & \textbf{97.7} & \textbf{91.7} & \textbf{89.1} & 83.7 & \textbf{98.0} & \textbf{97.4} & 90.7 & 84.2 & 97.5 & \textbf{94.7} & \textbf{94.4} & 39.4 & 88.2 $\green{(\uparrow 1.4)}$\\
\quad w/ \colearnplus (w/ CLIP)             &        & 97.4 & 89.4 & 88.0 & \textbf{86.0} & \textbf{98.0} & 96.4 & \textbf{93.9} & \textbf{85.2} & \underline{\textbf{97.8}} & 94.5 & 94.3 & \textbf{45.0} & \textbf{88.8} $\green{(\uparrow 2.0)}$\\ \midrule
NRC$^\dagger$ \citep{Yang2021ExploitingTI}   & \cmark & 96.8 & \textbf{92.0} & 83.8 & 57.2 & 96.6 & 95.3 & 84.2 & 79.6 & \textbf{94.3} & 93.9 & 90.0 & 59.8 & 85.3  \\
\quad w/ Co-learn (w/ ResNet-101)           &        & 96.9 & 89.2 & 81.1 & 65.5 & 96.3 & 96.1 & \textbf{89.8} & 80.6 & 93.7 & \textbf{95.4} & 88.8 & 60.0 & 86.1 $\green{(\uparrow 0.8)}$\\
\quad w/ Co-learn (w/ Swin-B)               &        & \textbf{97.4} & 91.3 & \textbf{84.5} & \textbf{65.8} & \textbf{96.9} & \textbf{97.6} & 88.8 & \textbf{82.0} & 93.8 & 94.7 & \textbf{91.1} & \textbf{61.6} & \textbf{87.1} $\green{(\uparrow 1.8)}$\\ \hdashline
\quad w/ Co-learn (w/ CLIP)                 &        & 97.5 & \textbf{91.9} & 83.7 & 65.0 & 96.7 & 97.5 & 88.3 & 81.1 & \textbf{93.0} & 95.5 & 91.6 & 59.5 & 86.8 $\green{(\uparrow 1.5)}$\\
\quad w/ \colearnplus (w/ CLIP)             &        & \textbf{98.0} & 90.8  & \textbf{83.9} & \textbf{69.0} & \textbf{97.4} & \textbf{97.6} & \textbf{91.7} & \textbf{81.6} & 92.8 & \textbf{96.2} & \textbf{92.8} & \textbf{59.9} & \textbf{87.6} $\green{(\uparrow 2.3)}$\\ \midrule
AaD$^\dagger$ \citep{Yang2022AttractingAD}   & \cmark & 96.9 & \textbf{90.2} & \textbf{85.7} & 82.8 & 97.4 & 96.0 & 89.7 & 83.2 & \textbf{96.8} & 94.4 & 90.8 & 49.0 & 87.7 \\
\quad w/ Co-learn (w/ ResNet-101)           &        & \textbf{97.7} & 87.9 & 84.8 & 79.6 & \textbf{97.6} & \textbf{97.5} & \textbf{92.4} & 83.7 & 95.3 & 94.2 & 90.3 & \textbf{57.4} & 88.2 $\green{(\uparrow 0.5)}$\\
\quad w/ Co-learn (w/ Swin-B)               &        & 97.6 & \textbf{90.2} & 85.0 & \textbf{83.1} & \textbf{97.6} & 97.1 & 92.1 & \textbf{84.9} & \textbf{96.8} & \textbf{95.1} & \textbf{92.2} & 56.8 & \textbf{89.1} $\green{(\uparrow 1.4)}$\\ \hdashline
\quad w/ Co-learn (w/ CLIP)                 &        & 97.5 & 91.4 & 85.4 & 82.4 & 97.3 & 97.8 & 92.3 & 81.7 & \textbf{95.7} & 94.3 & 92.5 & 51.7 & 88.3 $\green{(\uparrow 0.6)}$\\
\quad w/ \colearnplus (w/ CLIP)             &        & \textbf{97.7} & \textbf{92.1} & \textbf{87.1} & \textbf{83.5} & \textbf{98.1} & \textbf{98.3} & \textbf{93.7} & \textbf{85.8} & 95.4 & \textbf{95.6} & \textbf{94.0} & \textbf{64.0} & \textbf{90.4} $\green{(\uparrow 2.7)}$\\
\midrule[0.3pt]\bottomrule[1pt]
\end{tabular}
\end{adjustbox}
\caption{VisDA-C: 12-class classification accuracy of adapted ResNet-101. The source model is initialized with ImageNet-1k ResNet-101 weights. For proposed strategy, the pre-trained network used for co-learning is given in parenthesis: CLIP is pre-trained on WIT, and the rest are pre-trained on ImageNet-1k (i.e. no new data is introduced). CLIP Zero-shot utilizes the text template-based classifier from \citep{lin2023multimodality}. SF denotes source-free. $\dagger$ denotes reproduced results. \label{tab: visda_results}}
\end{table*}

\begin{table*}[tb]
\centering
 \setlength{\tabcolsep}{2pt}
\begin{adjustbox}{max width=0.95\textwidth}
\begin{tabular}{l*{13}{c}l}
\toprule[1pt]\midrule[0.3pt]

\textbf{Method}                 & \textbf{SF} & $\mathbf{C\rightarrow P}$ & $\mathbf{C\rightarrow R}$ & $\mathbf{C\rightarrow S}$  
                                & $\mathbf{P\rightarrow C}$ & $\mathbf{P\rightarrow R}$ & $\mathbf{P\rightarrow S}$ 
                                & $\mathbf{R\rightarrow C}$ & $\mathbf{R\rightarrow P}$ & $\mathbf{R\rightarrow S}$
                                & $\mathbf{S\rightarrow C}$ & $\mathbf{S\rightarrow P}$ & $\mathbf{S\rightarrow R}$ & \textbf{Avg}\\ \midrule
SHOT$^\dagger$ \cite{liang2020shot}         & \cmark & 62.0 & 78.0 & 59.9 & 63.9 & 78.8 & 57.4 & 68.7 & 67.8 & 57.7 & 71.5 & 65.5 & 76.3 & 67.3 \\
SHOT++$^\dagger$ \cite{liang2021shotplus}   & \cmark & \textbf{63.0} & \textbf{80.4} & \textbf{62.5} & \textbf{66.9} & \textbf{80.0} & 60.0 & \textbf{71.2} & \textbf{68.7} & \textbf{61.2} & 72.8 & 66.6 & \textbf{78.1} & \textbf{69.3} \\
AaD$^\dagger$ \cite{Yang2022AttractingAD}   & \cmark & 62.5 & 78.7 & 59.4 & 65.3 & 79.9 & \textbf{61.3} & 69.3 & 68.6 & 57.1 & \textbf{72.9} & \textbf{67.5} & 77.4 & 68.3 \\ \midrule
Source Only                     & \cmark & 47.2 & 61.7 & 50.7 & 47.2 & 71.7 & 44.3 & 58.2 & 63.2 & 49.2 & 59.5 & 52.4 & 60.5 & 55.5 \\
Co-learn (w/ ResNet-50)         & \cmark & 58.7 & 75.7 & 51.9 & 54.9 & 76.6 & 46.1 & 61.4 & 63.7 & 49.1 & 65.0 & 62.7 & 76.7 & 61.9 \\ 
Co-learn (w/ Swin-B)            & \cmark & \textbf{69.1} & \textbf{85.0} & \textbf{61.3} & \textbf{68.7} & \textbf{87.3} & \textbf{60.6} & \textbf{70.3} & \textbf{71.5} & \textbf{59.5} & \textbf{70.1} & \textbf{72.9} & \textbf{85.2} & \textbf{71.8} \\ \hdashline
CLIP Zero-shot                  & \cmark & 88.8 & 93.6 & 88.3 & 89.4 & 93.6 & 88.3 & 89.4 & 88.8 & 88.3 & 89.4 & 88.8 & 93.6 & 90.0 \\
Co-learn (w/ CLIP)              & \cmark & 75.1 & 86.5 & 78.5 & 78.9 & 86.7 & 76.8 & 85.4 & 79.1 & 76.7 & 81.2 & 73.8 & 84.4 & 80.3 \\
\colearnplus (w/ CLIP)          & \cmark & \underline{\textbf{89.5}} & \underline{\textbf{93.9}} & \underline{\textbf{88.6}} & \underline{\textbf{90.0}} & \underline{\textbf{93.8}} & \underline{\textbf{88.7}} & \underline{\textbf{90.3}} & \underline{\textbf{89.4}} & \underline{\textbf{88.5}} & \underline{\textbf{90.1}} & \underline{\textbf{89.5}} & \underline{\textbf{93.9}} & \underline{\textbf{90.5}}\\
\midrule[0.3pt]\bottomrule[1pt]
\end{tabular}
\end{adjustbox}
\caption{DomainNet: 126-class classification accuracy of adapted ResNet-50. The source model is initialized with ImageNet-1k ResNet-50 weights. For proposed strategy, the pre-trained network used for co-learning is given in parenthesis: CLIP is pre-trained on WIT, and the rest are pre-trained on ImageNet-1k (i.e. no new data is introduced). CLIP Zero-shot utilizes the text template-based classifier from \citep{lin2023multimodality}. SF denotes source-free. $\dagger$ denotes reproduced results.} \label{tab: domainnet_results}
\end{table*}

\noindent\textbf{Implementation details.}
We follow the network architecture and training scheme in \citep{liang2020shot,liang2021shotplus} to train source models: Office-31, Office-Home and DomainNet use ResNet-50 and VisDA-C uses ResNet-101 initialized with ImageNet-1k weights for feature extractor plus a 2-layer linear classifier with weight normalization, trained on labeled source data.
For our proposed co-learning strategy, we experiment with CLIP pre-trained on WebImage Text (WIT) \citep{Radford2021clip} and the following ImageNet-1k vision encoders curated on Torch~\citep{torchmodels} and Hugging Face~\citep{huggingfacemodels} as co-learning networks: ResNet-50, ResNet-101, ConvNeXt-S, Swin-S, ConvNeXt-B and Swin-B, where S and B denote the small and base versions of the architectures, respectively.
This list is not exhaustive and is meant as a demonstration that state-of-the-art networks can be successfully plugged into our proposed framework. We use ViT-L/14@336 CLIP model \citep{Radford2021clip} with a ViT transformer-based vision encoder. ConvNeXt convolutional networks~\citep{liu2022convnet} and Swin transformers~\citep{liu2021Swin} are recently-released architectures for computer vision tasks that demonstrated improved robustness to domain shifts~\citep{kim2022broad}. During target adaptation, we train using SGD optimizer for 15 episodes, with batch size 50 and learning rate 0.01 decayed to 0.001 after 10 episodes. We set confidence threshold $\gamma=0.5$ for Office-31, Office-Home and DomainNet and $\gamma=0.1$ for VisDA-C, with further analysis in Section~\ref{sec: further analysis}. For \colearnplus, we apply weak zero-shot guidance on Office-31 and Office-Home and strong zero-shot guidance on VisDA-C and DomainNet. We also apply our strategy on existing methods. For SHOT~\citep{liang2020shot} and SHOT++~\citep{liang2021shotplus} with pseudolabeling components, we replace the pseudolabels used with co-learned pseudolabels. For NRC~\citep{Yang2021ExploitingTI} and AaD~\citep{Yang2022AttractingAD} originally without pseudolabeling components, we add $0.3 L_{co-learning}$ to the training objective where the coefficient $0.3$ follows that in SHOT~\citep{liang2020shot} and SHOT++~\citep{liang2021shotplus}.

\subsection{Results}
\label{subsec: results}

We report results for co-learning with the ImageNet-1k network (ResNet-50 or ResNet-101) used for source model initialization, Swin-B transformer and vision-language CLIP in Table~\ref{tab: office31_results}, \ref{tab: officehome_results} and \ref{tab: visda_results}. In these experiments, co-learning with an ImageNet-1k network does not introduce any new data into the SFDA process, and further, co-learning with the ImageNet-1k network used for source model initialization does not introduce any new network architecture and hence feature extraction capabilities into the SFDA process. Best performance in each set of comparison is \textbf{bolded}, and best performance for each source-target transfer is \underline{underlined}. Full results on other architectures are in Appendix.

\noindent\textbf{Reusing pre-trained network from source model initialization:} Reusing the same ImageNet network from source model initialization for co-learning can improve the original source model performance. In this setup, no new data or network architecture has been introduced into the SFDA process, hence the improved performance demonstrates that co-learning can help to restore target information lost due to source training. Since Office-31 domains have realistic images of objects similar to ImageNet, the application of co-learning to SHOT and SHOT++ has no discernible impact on performance. NRC and AaD performance increase by $0.8\%$ and $0.6\%$, respectively. Several Office-Home and VisDA-C domains are characterized by non-realistic styles (e.g. Art, Clipart, Synthetic), and benefit more from co-learning with ImageNet networks. In these cases, existing methods demonstrate performance improvements of $0.5-0.9\%$ on Office-Home, and $0.5-1.7\%$ on VisDA-C.

\noindent\textbf{Co-learning with more robust pre-trained network:}
Co-learning with the more powerful and more robust ImageNet Swin-B significantly boosts performance in all evaluated setups. Note that no new data has been introduced into the SFDA process. Co-learn alone achieves a target accuracy of 93.6\% on Office-31, 83.5\% on Office-Home, 88.2\% on VisDA-C, and 71.8\% on DomainNet, beating even domain adaptation methods with joint source and target training. By incorporating Co-learn pseudolabels, existing SFDA methods improve by $2.0-3.3\%$ on Office-31, $3.6-9.4\%$ on Office-Home, and $1.4-2.8\%$ on VisDA-C.
Interestingly, on Office-31 and Office-Home, co-learning with ImageNet Swin-B alone is superior to integrating it with the existing SFDA methods tested, which have self-supervised learning components besides pseudolabeling. This observation suggests that effective target domain information in source models is limited, such that over-reliance on these models can impede target adaptation. In contrast, powerful and robust pre-trained networks can possess features that are more class-discriminative on target domains, and the substantial performance improvements underscore the advantage of using them in adapting source models.

\noindent\textbf{Co-learning with vision-language CLIP model:}
Compared to ImageNet vision models, CLIP is pre-trained with a larger and more diverse multimodal dataset to connect visual and textual concepts. Co-learning with just the CLIP vision encoder significantly boosts performance, and leveraging CLIP's text encoder and zero-shot classification capabilities further increases performance. Co-learn achieves a target accuracy of 94.4\% on Office-31, 84.0\% on Office-Home, 88.6\% on VisDA-C, and 80.3\% on DomainNet. This performance is $0.4-8.5\%$ higher than that of Co-learn with ImageNet Swin-B. By leveraging CLIP's zero-shot classification capabilities, \colearnplus\ achieves a target accuracy of 94.8\% on Office-31, 87.1\% on Office-Home, 91.1\% on VisDA-C, and 90.5\% on DomainNet, which is $0.4-10.2\%$ higher than Co-learn performance across these 4 datsets. Co-learning with CLIP also improves existing SFDA methods.
By incorporating Co-learn pseudolabels, existing methods improve by $1.7-2.7\%$ on Office-31, $3.1-10.0\%$ on Office-Home, and $0.6-3.4\%$ on VisDA-C. By incorporating \colearnplus\ pseudolabels, existing methods improve by $1.4-3.3\%$ on Office-31, $3.4-12.2\%$ on Office-Home, and $2.0-4.4\%$ on VisDA-C.

We note that source model co-learned with CLIP outperforms existing SFDA methods incorporated with co-learning. Source model with co-learning relies solely on the pseudo-labels produced collaboratively by the source model and the co-learning model. We expect the resulting adaptation performance to improve with stronger co-learning models, as demonstrated in detail in Table~\ref{tab: results_full} in the Appendix. Existing SFDA methods have other training objectives to mine target data structures in the source model. When the co-learning model is much stronger than the source model (e.g. CLIP vs ResNet-50), the objectives of mining target structures in the source model may be suboptimal and hinder learning. Tuning hyperparameters for each SFDA method to balance the multiple objectives may produce better results but is out-of-scope of this work, as our aim is to demonstrate that co-learning can improve existing SFDA methods.

Since co-learning involves processing outputs from two models, it can incur longer training time. For adaptation from Amazon to DSLR domain in Office dataset, SHOT takes 124s, SHOT++ takes 1049s, AaD takes 353s, and NRC takes 179s, following training schemes set in the original papers. Co-learn takes 730s and \colearnplus\ takes 899s for 15 epochs of training. However, we note that Co-learn and \colearnplus\ converge after 3 and 5 epochs, respectively, meaning that the full 15 epochs are not needed to achieve optimal adaptation performance. The pre-trained model is discarded after training, hence the co-learned model incur the same inference cost as other adapted models.

\section{Further Analysis}
\label{sec: further analysis}

We conduct further experiments with our proposed strategy in Section~\ref{subsec: further_analysis_colearn}, \ref{subsec: further_analysis_colearnplus} and \ref{subsec: other_adaptation_scenarios}.

\subsection{Two-branch framework}
\label{subsec: further_analysis_colearn}

We fix the pre-trained model branch to use ImageNet-1k ConvNeXt-S, which has an intermediate parameter size in our networks tested.

\noindent\textbf{Adaptation model branch.}
We experiment with alternative pseudolabeling strategies besides MatchOrConf with a subset of domain pairs from the Office-Home dataset in Table~\ref{tab: ablation_pseudolabel_strategy}: SelfConf selects confident samples from the adaptation model branch, OtherConf selects confident samples from the pre-trained model branch, Match selects samples with the same predictions on both branches regardless of confidence, and MatchAndConf selects confident samples with the same predictions on both branches. SelfConf has the worst performance as source model confidence is not well-calibrated on target domain. Overall, MatchOrConf is the best strategy. From Table~\ref{tab: ablation_confidence}, the optimal confidence threshold $\gamma$ differs across datasets. We estimate the target-compatibiilty ratio of the source to pre-trained feature extractor using the ratio of their oracle target domain accuracy. We compute the oracle target domain accuracy by fitting a nearest-centroid-classification head using fully-labeled target samples on top of the feature extractor. When the ratio is low as in VisDA-C, the pre-trained ImageNet feature extractor is more target-compatible, and lowering $\gamma$ facilitates the pseudolabeling of more samples based on its predictions. In the absence of labeled target labels, we estimate the ratio using CLIP zero-shot predictions and obtain similar observations.
In practice, we set $\gamma=0.1$ for VisDA-C as the Real target domain is more similar to ImageNet than to the Synthetic source even by visual inspection, and default to $\gamma=0.5$ otherwise although it may not be the optimal value for other datasets.

\begin{table}

\begin{subtable}[t]{0.48\textwidth}
\centering
 \setlength{\tabcolsep}{2pt}
\begin{adjustbox}{max width=0.85\columnwidth}
\begin{tabular}{l*{4}{c}}
\toprule[1pt]\midrule[0.3pt]
\textbf{Pseudolabel strategy}   & $\mathbf{A\rightarrow C}$ & $\mathbf{A\rightarrow P}$ & $\mathbf{A\rightarrow R}$ & \textbf{Avg} \\ \midrule
SelfConf                        & 49.1  & 74.8  & 77.1  & 67.0 \\
OtherConf                       & 57.5  & \emph{85.4}  & \emph{86.7}  & 76.5 \\
Match                           & \textbf{61.3}  & 84.2  & 86.0  & \emph{77.2} \\
MatchOrConf                     & 59.7  & \textbf{86.3}  & \textbf{87.1}  & \textbf{77.7} \\
MatchAndConf                    & \emph{60.3}  & 81.3  & 84.5  & 75.4 \\
\midrule[0.3pt]\bottomrule[1pt]
\end{tabular}
\end{adjustbox}
\caption{Pseudolabeling strategies, on Office-Home \label{tab: ablation_pseudolabel_strategy}}
\end{subtable}

\begin{subtable}[t]{0.48\textwidth}
\centering
 \setlength{\tabcolsep}{2pt}
\begin{adjustbox}{max width=\columnwidth}
\begin{tabular}{lP{1.5cm}P{2cm}P{1.4cm}}
\toprule[1pt]\midrule[0.3pt]
\textbf{Confidence threshold $\gamma$}    & \textbf{Office-31}   & \textbf{Office-Home}  & \textbf{VisDA-C} \\ \midrule
0.1 & 90.5              & 76.1              & \textbf{87.1} \\
0.3 & \textbf{91.2}     & 77.1              & \emph{86.8} \\
0.5 & \emph{91.0}  & \emph{77.4}  & 86.4 \\
0.7 & 90.8              & \textbf{77.5}     & 85.8 \\
0.9 & 90.8              & 77.3              & 85.5 \\ \midrule
Target-compatibility ratio  & 0.982 & 0.947 & 0.844 \\
CLIP-estimated ratio  & 0.944 & 0.879 & 0.795 \\
\midrule[0.3pt]\bottomrule[1pt]
\end{tabular}
\end{adjustbox}
\caption{MatchOrConf confidence threshold $\gamma$. The target-compatibility ratio of source to pre-trained model oracle target accuracy suggests a lower $\gamma$ when the ratio is low. The estimated ratio using CLIP zero-shot predictions suggests similarly. \label{tab: ablation_confidence}}
\end{subtable}

\caption{Co-learning experiments with ImageNet-1k ConvNeXt-S in pre-trained model branch. \label{tab: ablation_adaptation_model_branch}}
\end{table}

\noindent\textbf{Pre-trained model branch.}
We conduct experiments exploring different updates to the pre-trained model branch using a subset of domain pairs from the Office-Home dataset in Table~\ref{tab: ablation_pretrained_model_branch}. Specifically, we consider updating either no component or different component(s) of the pre-trained network, including the feature extractor, weighted nearest-centroid-classifier (NCC) and a linear 1-layer logit projection layer inserted after NCC. Finetuning just the NCC classifier yields the best result overall. The effectiveness of co-learning depends on the two branches offering different views on classification decisions. Finetuning the feature extractor or projection layer runs the risk of the pre-trained model predictions converging too quickly to the adaptation model predictions.

\begin{table}[tb]

\centering
 \setlength{\tabcolsep}{2pt}
\begin{adjustbox}{max width=\columnwidth}
\begin{tabular}{P{1.6cm}P{1.5cm}P{1.5cm}*{4}{c}}
\toprule[1pt]\midrule[0.3pt]
\textbf{FE} & \textbf{CLF}    & \textbf{Projection} & $\mathbf{A\rightarrow C}$ & $\mathbf{A\rightarrow P}$ & $\mathbf{A\rightarrow R}$ & \textbf{Avg} \\ \midrule
\xmark      & \xmark        & \xmark                & 59.4  & 83.8  & 86.5  & 76.6 \\
\checkmark  & \xmark        & \xmark                & 59.6  & 82.6  & 86.4  & 76.2 \\
\checkmark  & \checkmark    & \xmark                & \textbf{60.3}  & 84.9  & 86.4  & 77.2 \\
\xmark      & \checkmark    & \xmark                & \emph{59.7}  & \textbf{86.3}  & \emph{87.1}  & \textbf{77.7} \\
\xmark      & \checkmark    & \checkmark            & 59.5  & \emph{85.9}  & \textbf{87.2}  & \emph{77.5} \\
\xmark      & \xmark        & \checkmark            & 59.5  & 84.1  & 86.5  & 76.7 \\
\midrule[0.3pt]\bottomrule[1pt]
\end{tabular}
\end{adjustbox}

\caption{Co-learning experiments on the component finetuned in the ImageNet-1k ConvNeXt-S pre-trained model branch, on Office-Home. Components include the feature extractor (FE), weighted nearest-centroid-classifier (CLF), and a linear projection layer inserted after the classifier. \label{tab: ablation_pretrained_model_branch}}
\end{table}

\begin{figure}
  \centering
  \begin{subfigure}{0.45\linewidth}
    \includegraphics[width=\linewidth]{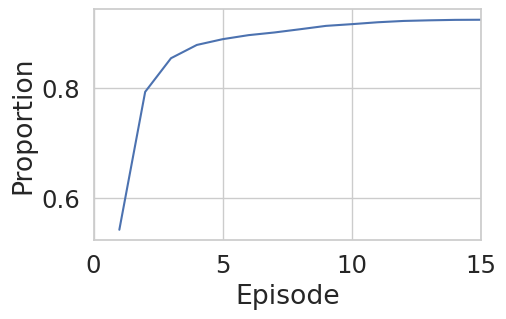}
    \caption{Proportion of pseudolabels}
    \label{fig: pseudolabel_proportion}
  \end{subfigure}
  \hfill
  \begin{subfigure}{0.45\linewidth}
    \includegraphics[width=\linewidth]{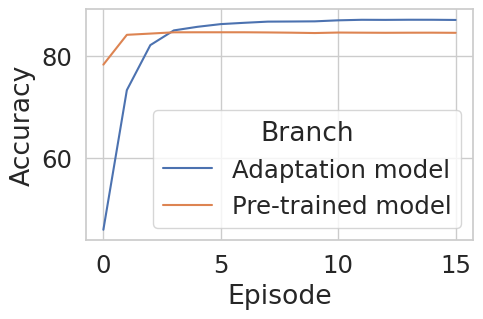}
    \caption{Classification accuracy}
    \label{fig: two_branch_accuracy}
  \end{subfigure}
  
  \caption{VisDA-C co-learning training curves, with ImageNet-1k ConvNeXt-S in pre-trained model branch.}
  \label{fig: training_curves}
\end{figure}

\noindent\textbf{Training curves.} Figure~\ref{fig: training_curves} visualizes the co-learning process for VisDA-C. The proportion of target pseudolabels rises from 0.542 to 0.925 over 15 episodes. Classification accuracy on the pre-trained model branch starts at a higher level as the ImageNet feature extractor is more target-compatible than the source feature extractor is. However, its accuracy curve levels off earlier as the pre-trained feature extractor is fixed. The adaptation model learns from the pre-trained model predictions, gradully surpassing its accuracy as the adaptation network feature extractor continues to adapt.

\subsection{Integrating zero-shot CLIP}
\label{subsec: further_analysis_colearnplus}

An important element in integrating a pre-trained network into the co-learning framework is designing the task-specifc classification head to be fitted on top of the pre-trained feature extractor $f_*$. The classification head affects the accuracy of the pre-trained network predictions on the target domain, and consequently the quality of generated pseudolabels for adapting the source model. A specific consideration for CLIP is how to leverage its zero-shot capabilities. In Table~\ref{tab: clip_zero_shot}, we explore different formulations of the classification head and report their corresponding target accuracy. The zero-shot classifier \citep{lin2023multimodality} detailed in Equation~\ref{eqn: text_pretrained_centroid}-\ref{eqn: text_pretrained_prob} is a nearest-centroid-classifier (NCC) where the class centroids are derived from text embeddings of the class labels. The Co-learn classifier $q_*$ defined in Equation~\ref{eqn: pretrained_centroid}-\ref{eqn: pretrained_prob} is a weighted NCC with the weights derived from source model predictions. However, neither classifier is optimal, as they only consider either the text space or visual space. In contrast, the \colearnplus\ classifier $q^{\plus}_*$ defined in Equation~\ref{eqn: clip_pretrained_centroid}-\ref{eqn: clip_pretrained_prob} effectively integrates classification decisions in both the text and visual space.

The strength of CLIP's zero-shot guidance applied in \colearnplus\ depends on the quality of the text-classifier. Although CLIP is trained for image-text alignment, distribution shift is still observed between text and image embeddings \citep{tanwisuth2023pouf, xuefeng2023reclip}, such that over-relying on the text-classifier can be sub-optimal. In Table~\ref{tab: clip_guidance}, we compare the quality of the zero-shot text-classifier and the image-classifier applied in weak guidance defined in Equation~\ref{eqn: clip_pretrained_centroid}-\ref{eqn: clip_pretrained_prob} with $\alpha = 1$, constructed as a weighted nearest-centroid-classification head fit on image embeddings with text-classifier predictions. 
We measure the quality of each classifier with a small number of target labeled samples. We compute 3-shot accuracy, and report the average over 3 seeds in Table~\ref{tab: clip_guidance}. For VisDA-C which is evaluated by macro-average accuracy across classes, we randomly select 3 samples per class. For the other datasets which are evaluated by micro-average accuracy across all samples, we randomly select ($3 \times \# classes$) samples. From Table~\ref{tab: clip_guidance}, for Office-31 and Office-Home, the image-classifier applied in weak guidance has better few-shot accuracy than the text-classifier. For VisDA-C and DomainNet, the text-classifier has better few-shot accuracy, and we apply strong guidance by setting  $\alpha = 0.5$ as in an ensemble classifier. Detailed results for each domain pair are provided in Appendix \ref{appendix: Integrating zero-shot CLIP}.

\begin{table}[tb]

\centering
 \setlength{\tabcolsep}{2pt}
\begin{adjustbox}{max width=0.88\columnwidth}
\begin{tabular}{lP{1.5cm}P{2cm}P{1.4cm}}
\toprule[1pt]\midrule[0.3pt]
\textbf{CLF}        & \textbf{Office-31} & \textbf{Office-Home} & \textbf{VisDA-C} \\ \midrule
Zero-shot                    & 87.5 & 81.5 & 90.2\\
Co-learn $q_*$      & 92.4 & 81.4 & 75.3 \\
\colearnplus\ $q^{\plus}_*$ & \textbf{95.0} & \textbf{86.7} & \textbf{91.1} \\
\midrule[0.3pt]\bottomrule[1pt]
\end{tabular}
\end{adjustbox}

\caption{Classification accuracy of CLIP vision encoder + classification head (CLF). Zero-shot classifier follows \citep{lin2023multimodality}, Co-learn and \colearnplus\ classifier is the weighted nearest-centroid-classifier initialized at the start of target adaptation. \label{tab: clip_zero_shot}}
\end{table}

\begin{table}[tb]

\centering
 \setlength{\tabcolsep}{2pt}
\begin{adjustbox}{max width=\columnwidth}
\begin{tabular}{lP{1.4cm}P{1.9cm}P{1.4cm}P{1.7cm}}
\toprule[1pt]\midrule[0.3pt]
\textbf{Guidance}        & \textbf{Office-31} & \textbf{Office-Home} & \textbf{VisDA-C} & \textbf{DomainNet}\\ \midrule
weak        & \textbf{94.8} & \textbf{87.1} & 89.3 & 86.8 \\
strong      & 92.5 & 83.7 & \textbf{91.1} & \textbf{90.5}\\ \midrule
image-clf@3 & 90.6 & 81.3 & 86.1 & 88.5 \\
text-clf@3  & 88.4 & 81.0 & 88.9 & 90.2 \\
ratio       & $> 1$ & $> 1$ & $< 1$ & $< 1$ \\
\midrule[0.3pt]\bottomrule[1pt]
\end{tabular}
\end{adjustbox}

\caption{Comparison of \colearnplus\ classification accuracy with different strength of CLIP's text-classifier-based zero-shot guidance. The values `image-clf@3' and `text-clf@3' measure the 3-shot target domain accuracy of the image-classifier and text-classifier, respectively.
Weak guidance is preferred when the text-classifier results in worse prediction. \label{tab: clip_guidance}}
\end{table}

\subsection{Other adaptation scenarios}
\label{subsec: other_adaptation_scenarios}

\noindent \textbf{Non-closed-set settings.}
Our co-learning strategy is effective even in the presence of label shift where source and target label spaces do not match, as shown in Table~\ref{tab: openpartial_officehome_results} on the Office-Home dataset. In the open-set scenario, the first 25 classes are target private classes, and the next 40 are shared classes. In the partial-set scenario, the first 25 classes are shared classes, and the next 40 are source private classes. In the open-partial scenario, the first 10 classes are shared classes, the next 5 are source private classes, and the remaining 50 are target private classes, and we report the harmonic mean of known and unknown class accuracy (H-score) following \citep{yang2022onering}. We add cross-entropy loss with co-learned pseudolabels at temperature 0.1 on the open-set and partial-set version of SHOT and the open-partial method OneRing \citep{yang2022onering}. From Table~\ref{tab: openpartial_officehome_results}, the performance improvement is $0.9-2.4\%$ in the open-set scenario, $0.3-2.0\%$ in the partial-set scenario, and $0.7-0.9\%$ in the open-partial scenario.

\noindent \textbf{Multi-source adaptation.}
Our proposed strategy can work in multi-source SFDA as well. In Table~\ref{tab: multisource_results} on Office-31, we incorporate co-learned pseudolabels into the multi-source method CAiDA \citep{dong2021caida} and improves its performance by $0.5-3.5\%$.

\begin{table*}[tb]
\centering
 \setlength{\tabcolsep}{2pt}
\begin{adjustbox}{max width=\textwidth}
\begin{tabular}{l*{12}{c}l}
\toprule[1pt]\midrule[0.3pt]
\textbf{Method}                 & $\mathbf{A\rightarrow C}$ & $\mathbf{A\rightarrow P}$ & $\mathbf{A\rightarrow R}$  
                                & $\mathbf{C\rightarrow A}$ & $\mathbf{C\rightarrow P}$ & $\mathbf{C\rightarrow R}$ 
                                & $\mathbf{P\rightarrow A}$ & $\mathbf{P\rightarrow C}$ & $\mathbf{P\rightarrow R}$
                                & $\mathbf{R\rightarrow A}$ & $\mathbf{R\rightarrow C}$ & $\mathbf{R\rightarrow P}$ & \textbf{Avg}\\ \midrule
Open-set SHOT$^\dagger$ \citep{liang2020shot}          & 52.8 & 74.5 & 77.1 & \textbf{57.6} & 71.8 & 74.2 & 51.2 & 45.0 & 76.8 & 61.8 & 57.2 & 81.7 & 65.1 \\
\quad w/ Co-learn (w/ ResNet-50)            & \textbf{55.0} & 76.1 & 78.2 & 55.0 & \textbf{73.7} & 73.7 & 53.2 & 47.6 & 77.7 & 61.8 & 57.5 & 82.2 & 66.0 $\green{(\uparrow 0.9)}$ \\
\quad w/ Co-learn (w/ Swin-B)               & 52.2 & \textbf{77.4} & \textbf{78.6} & 56.3 & 71.3 & \textbf{75.3} & \textbf{53.9} & \textbf{49.9} & \textbf{77.9} & \textbf{62.5} & \textbf{58.2} & \textbf{82.3} & \textbf{66.3} $\green{(\uparrow 1.2)}$\\ \hdashline
\quad w/ Co-learn (w/ CLIP)                 & \textbf{55.0} & 76.8 & \textbf{78.5} & \textbf{60.9} & \textbf{74.5} & 74.4 & 55.6 & 48.7 & \textbf{78.0} & 63.0 & \textbf{59.8} & \textbf{85.3} & \textbf{67.5} $\green{(\uparrow 2.4)}$\\
\quad w/ \colearnplus (w/ CLIP)             & 54.9 & \textbf{77.6} & 78.4 & 60.4 & 72.7 & \textbf{75.9} & \textbf{56.0} & \textbf{51.0} & 77.2 & \textbf{64.0} & 58.8 & 83.6 & \textbf{67.5} $\green{(\uparrow 2.4)}$\\ \midrule

Partial-set SHOT$^\dagger$ \citep{liang2020shot}       &  65.9 & \textbf{86.1} & 91.4 & 74.6 & 73.6 & 85.1 & 77.3 & 62.9 & 90.3 & 81.8 & 64.9 & 85.6 & 78.3 \\
\quad w/ Co-learn (w/ ResNet-50)            &  63.9 & 84.7 & 91.8 & 77.6 & 73.6 & 84.5 & 77.4 & 64.8 & \textbf{90.5} & 81.4 & 65.1 & \textbf{87.8} & 78.6 $\green{(\uparrow 0.3)}$ \\
\quad w/ Co-learn (w/ Swin-B)               &  \textbf{66.2} & 85.9 & \textbf{93.2} & \textbf{78.2} & \textbf{74.7} & \textbf{85.6} & \textbf{81.5} & \textbf{65.8} & 90.4 & \textbf{82.6} & \textbf{65.4} & 87.7 & \textbf{79.8} $\green{(\uparrow 1.5)}$ \\ \hdashline
\quad w/ Co-learn (w/ CLIP)                 & 67.0 & \textbf{87.3} & \textbf{91.8} & 77.0 & 71.2 & 85.2 & 77.3 & \textbf{70.2} & \textbf{91.3} & \textbf{84.0} & \textbf{71.3} & \textbf{87.8} & 80.1 $\green{(\uparrow 1.8)}$ \\
\quad w/ \colearnplus (w/ CLIP)             & \textbf{68.0} & 86.6 & 91.7 & \textbf{77.7} & \textbf{73.1} & \textbf{89.3} & \textbf{79.2} & 68.3 & 91.2 & 83.3 & 68.2 & 87.5 & \textbf{80.3} $\green{(\uparrow 2.0)}$\\ \midrule

Open-partial OneRing$^\dagger$ \citep{yang2022onering} & \textbf{68.1} & 81.9 & 87.6 & 72.2 & 76.9 & 83.3 & 80.7 & \textbf{68.8} & 88.2 & 80.5 & 66.0 & 85.5 & 78.3 \\ 
\quad w/ Co-learn (w/ ResNet-50)            & 64.9 & \textbf{86.8} & \textbf{88.0} & 71.9 & 76.7 & 83.5 & \textbf{82.2} & 67.6 & \textbf{89.2} & 82.8 & \textbf{70.0} & 86.4 & \textbf{79.2} $\green{(\uparrow 0.9)}$ \\
\quad w/ Co-learn (w/ Swin-B)               & 65.8 & 85.9 & \textbf{88.0} & \textbf{72.8} & \textbf{77.9} & \textbf{83.9} & 81.9 & 65.8 & 88.5 & \textbf{82.9} & 67.2 & \textbf{87.1} & 79.0 $\green{(\uparrow 0.7)}$ \\ \hdashline
\quad w/ Co-learn (w/ CLIP)                 & \textbf{65.0} & \textbf{86.9} & \textbf{87.5} & 73.7 & 77.3 & \textbf{83.5} & \textbf{81.6} & \textbf{67.2} & \textbf{89.3} & 81.7 & 69.5 & \textbf{86.5} & \textbf{79.1} $\green{(\uparrow 0.8)}$\\
\quad w/ \colearnplus (w/ CLIP)             & 63.4 & 86.6 & 87.3 & \textbf{74.1} & \textbf{77.9} & 83.3 & 81.5 & 66.6 & \textbf{89.3} & \textbf{82.4} & \textbf{70.0} & 86.2 & 79.0 $\green{(\uparrow 0.7)}$\\
\midrule[0.3pt]\bottomrule[1pt]
\end{tabular}
\end{adjustbox}
\caption{Office-Home: Accuracy for open-set and partial-set setting, H-score for open-partial setting, of adapted ResNet-50. $\dagger$ denotes reproduced results according to our experiment setups. \label{tab: openpartial_officehome_results}}
\end{table*}

\begin{table}[tb]
\centering
 \setlength{\tabcolsep}{2pt}
\begin{adjustbox}{max width=0.92\columnwidth}
\begin{tabular}{l*{3}{c}l}
\toprule[1pt]\midrule[0.3pt]
\textbf{Method}                 & $\mathbf{\rightarrow A}$ & $\mathbf{\rightarrow D}$ & $\mathbf{\rightarrow W}$ 
                                & \textbf{Avg}\\ \midrule                                
CAiDA$^\dagger$ \citep{dong2021caida}                  & 75.7 & 98.8 & 93.2 & 89.2  \\ 
\quad w/ Co-learn (w/ ResNet-50)            & 76.8 & 99.0 & 93.2 & 89.7 $\green{(\uparrow 0.5)}$ \\
\quad w/ Co-learn (w/ Swin-B)               & \textbf{79.3} & \textbf{99.6} & \textbf{97.4} & \textbf{92.1} $\green{(\uparrow 2.9)}$\\ \hdashline
\quad w/ Co-learn (w/ CLIP)                 & 80.1 & \textbf{99.6} & \textbf{97.7} & 92.5 $\green{(\uparrow 3.3)}$\\
\quad w/ \colearnplus (w/ CLIP)             & \textbf{80.9} & \textbf{99.6} & \textbf{97.7} & \textbf{92.7} $\green{(\uparrow 3.5)}$\\
\midrule[0.3pt]\bottomrule[1pt]
\end{tabular}
\end{adjustbox}
\caption{Multi-source Office-31 accuracy of adapted ResNet-50. $\dagger$ denotes reproduced results. \label{tab: multisource_results}}
\end{table}

\section{Discussion}
\label{sec: further_analysis_discussion}

We consider the characteristics that make pre-trained networks suitable for co-learning and further discuss the effectiveness of our proposed strategy.

\begin{figure}
  \centering
  \begin{subfigure}{0.49\linewidth}
    \includegraphics[width=\linewidth]{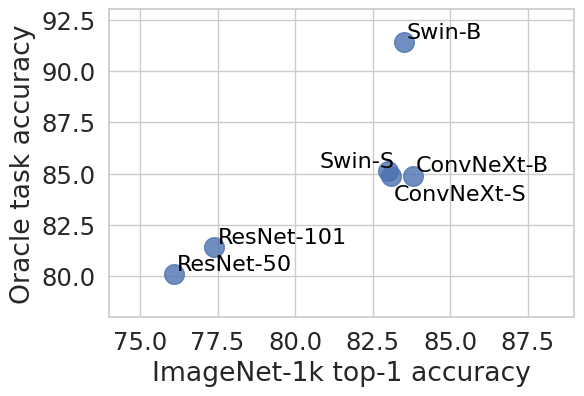}
    \caption{Oracle target accuracy versus ImageNet-1k top-1 accuracy}
  \end{subfigure}
  \hfill
  \begin{subfigure}{0.49\linewidth}
    \includegraphics[width=\linewidth]{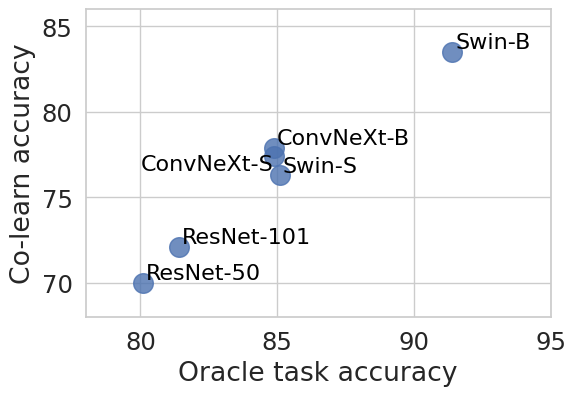}
    \caption{Accuracy after co-learning versus oracle target accuracy}
  \end{subfigure}
  \vspace{-2mm}
  \caption{Evaluations of ImageNet-1k networks. Oracle target accuracy of the ImageNet-1k feature extractors and accuracy of adapted model after co-learning with the ImageNet-1k feature extractors are assessed on Office-Home.}
  \label{fig: target_compatibility}
  \vspace{-4mm}
\end{figure}

\emph{What characteristics are preferred for the pre-trained feature extractor in co-learning?} 
We first study this question in Figure~\ref{fig: target_compatibility} through the relationship of ImageNet-1k top-1 accuracy and Office-Home oracle target accuracy. The oracle target accuracy is computed by fitting a nearest-centroid-classification head on the feature extractor using fully-labeled target samples. In general, a feature extractor with higher ImageNet accuracy has learned better representations and more robust input-to-feature mappings. Consequently, it is likely to be more transferable, have higher oracle accuracy on the new target domain, and hence be more helpful in adapting the source model through co-learning. 
Next, we analyze a few specific source-domain pairs in Table~\ref{tab: comparison_resnet} by plugging source and ImageNet-1k feature extractors into the pre-trained model branch. We observe the impacts of several positive characteristics: 
\begin{enumerate}
    \item Dataset similarity (inputs and task);
    \item Robustness against covariate shift;
    \item Provision of alternative views.
\end{enumerate}
(1) Dataset similarity (input and task): In $C\rightarrow R$ with the same ResNet-50 architecture, ImageNet samples exhibit more similarity to the Real World target domain than the Clipart source domain. The ImageNet-1k ResNet-50 has higher oracle target accuracy than Source ResNet-50. (2) Robustness against covariate shift: In $A\rightarrow C$, although ImageNet is less similar to the Clipart target domain than the Art source domain, replacing ResNet-50 with the more powerful ResNet-101 in the pre-trained model branch improves oracle target accuracy. (3) Provision of alternative views: In $R\rightarrow A$, although ImageNet-1k ResNet-50 has slightly lower oracle target accuracy than Source ResNet-50 ($81.2\%$ vs. $81.8\%$), co-learning with it results in better adaptation performance ($71.1\%$ vs. $70.5\%$). Since the adaptation model branch is already initialized by source model, utilizing an ImageNet feature extractor in the other branch provides an alternative view on features, and consequently classification decision, thereby benefiting co-learning. 

\begin{table}

\centering
\setlength{\tabcolsep}{2pt}
\begin{adjustbox}{max width=0.97\columnwidth}
\begin{tabular}{l*{4}{c}}
\toprule[1pt]\midrule[0.3pt]
\textbf{Model}  & $\mathbf{A\rightarrow C}$ & $\mathbf{A\rightarrow P}$ & $\mathbf{A\rightarrow R}$ & \textbf{Avg} \\ \midrule
ConvNext-B + classification head$^\dagger$  & 56.1  & 78.6  & 83.3  & 72.7 \\
ResNet-50 adapted w/ ConvNeXt-B             & \textbf{60.5}  & \textbf{86.2}  & \textbf{87.3}  & \textbf{78.0} \\ \midrule
Swin-B + classification head$^\dagger$      & 67.1  & 86.7  & 89.1  & 81.0 \\
ResNet-50 adapted w/ Swin-B                 & \textbf{69.6}  & \textbf{89.6}  & \textbf{91.2}  & \textbf{83.5}\\ \midrule
CLIP + classification head$^\dagger$  & 77.0  & 88.3  & 89.7  & 85.0 \\
ResNet-50 adapted w/ CLIP             & \textbf{80.0}  & \textbf{91.2}  & \textbf{91.8}  & \textbf{87.7} \\
\midrule[0.3pt]\bottomrule[1pt]
\end{tabular}
\end{adjustbox}

\caption{Comparison of classification accuracy of large-data pretrained feature extractor + source-trained classification head, versus classification accuracy of adapted ResNet-50, on Office-Home. $\dagger$ denotes classifier is trained on fully-labeled source data. \label{tab: comparison_two_branch}}
\end{table}

\begin{table*}[tb]
\centering
\begin{adjustbox}{max width=0.75\textwidth}
\begin{tabular}{l*{7}{c}l}
\toprule[1pt]\midrule[0.3pt]
\textbf{Method}                 & \textbf{\# Param} & $\mathbf{A\rightarrow D}$ & $\mathbf{A\rightarrow W}$ & $\mathbf{D\rightarrow A}$  
                                & $\mathbf{D\rightarrow W}$ & $\mathbf{W\rightarrow A}$ & $\mathbf{W\rightarrow D}$ &\textbf{Avg} \\ \midrule                   
POUF                      & 428M  & 98.2 & 98.2 & \textbf{87.7} & 98.2 & \textbf{87.7} & 98.2 & 94.7 \\
\colearnplus (w/ POUF)    & 25M   & \textbf{99.0} & \textbf{99.4} & 87.0 & \textbf{99.6} & 87.3 & \textbf{100.0} & \textbf{95.4} $\green{(\uparrow 0.7)}$\\ \hdashline
ReCLIP                    & 428M  & 86.9 & 87.7 & \textbf{84.1} & 87.7 & 84.1 & 86.9 & 86.2\\
\colearnplus (w/ ReCLIP)  & 25M   & \textbf{89.8} & \textbf{92.3} & 84.0 & \textbf{93.7} & \textbf{84.2} & 9\textbf{9.2} & \textbf{90.5} $\green{(\uparrow 4.3)}$\\ 
\midrule[0.3pt]\bottomrule[1pt]
\end{tabular}
\end{adjustbox}
\caption{Office-31: 31-class classification accuracy of adapted models. POUF and ReCLIP are CLIP models adapted using target datasets. $\dagger$ denotes reproduced results. \label{tab: office31_poufreclip}}
\end{table*}

\begin{table*}[tb]
\centering
\begin{adjustbox}{max width=\textwidth}
\begin{tabular}{l*{13}{c}l}
\toprule[1pt]\midrule[0.3pt]
\textbf{Method}                 & \textbf{\# Param} & $\mathbf{A\rightarrow C}$ & $\mathbf{A\rightarrow P}$ & $\mathbf{A\rightarrow R}$  
                                & $\mathbf{C\rightarrow A}$ & $\mathbf{C\rightarrow P}$ & $\mathbf{C\rightarrow R}$ 
                                & $\mathbf{P\rightarrow A}$ & $\mathbf{P\rightarrow C}$ & $\mathbf{P\rightarrow R}$
                                & $\mathbf{R\rightarrow A}$ & $\mathbf{R\rightarrow C}$ & $\mathbf{R\rightarrow P}$ & \textbf{Avg}\\ \midrule                 
POUF                      & 428M  & 83.6 & \textbf{95.8} & \textbf{94.5} & 90.1 & \textbf{95.8} & \textbf{94.5} & \textbf{90.1} & 83.6 & \textbf{94.5} & 90.2 & 83.6 & \textbf{95.8} & \textbf{91.0}\\
\colearnplus (w/ POUF)    & 25M   & \textbf{84.0} & 95.3 & 93.6 & \textbf{90.2} & 95.1 & 93.4 & 89.9 & \textbf{84.4} & 93.9 & \textbf{90.6} & \textbf{84.0} & 95.4 & 90.8 $\red{(\downarrow 0.2)}$\\ \hdashline
ReCLIP                    & 428M  & 79.1 & \textbf{94.3} & \textbf{94.1} & 87.7 & 94.3 & \textbf{94.1} & 87.7 & 79.1 & \textbf{94.1} & 87.7 & 79.1 & 94.3 & 88.8\\
\colearnplus (w/ ReCLIP)  & 25M   & \textbf{80.3} & 93.9 & 93.0 & \textbf{88.0} & \textbf{94.9} & 93.2 & \textbf{88.2} & \textbf{81.0} & 93.3 & \textbf{88.5} & \textbf{81.6} & \textbf{94.5} & \textbf{89.2} $\green{(\uparrow 0.4)}$\\ 
\midrule[0.3pt]\bottomrule[1pt]
\end{tabular}
\end{adjustbox}
\caption{Office-Home: 65-class classification accuracy of adapted models. POUF and ReCLIP are CLIP models adapted using target datasets. $\dagger$ denotes reproduced results. \label{tab: officehome_poufreclip}}
\end{table*}

\emph{Given that modern networks have learned more effective and robust representations, is it sufficient to use them and not adapt?} In Table~\ref{tab: comparison_two_branch}, we fit a 2-layer linear classification head as in Section~ \ref{subsec: experimental setups} on ImageNet-1k ConvNext-B and Swin-B feature extractors and WIT CLIP vision encoder by accessing and training the classifier on fully-labeled source data. On the Office-Home domain pairs tested, the average performance of CLIP + classification head is 4.0\% higher than Swin-B + classification head and 12.3\% higher than ConvNext-B + classification head, demonstrating that the choice of feature extractor can indeed mitigate the adverse effects of domain shift. However, adaptation to the target domain remains necessary. Without having to access the source data and utilizing a larger source model, the ResNet-50 adapted with our proposed strategy achieves 2.5-5.3\% higher average classification accuracy. Similarly, relying solely on zero-shot image recognition capabilities in CLIP is insufficient. From Table~\ref{tab: clip_zero_shot} and Table~\ref{tab: office31_results}-\ref{tab: visda_results}, the zero-shot CLIP accuracy can be significantly lower than that of the ResNets adapted with our proposed strategy. For instance, zero-shot CLIP accuracy is lower than \colearnplus\ accuracy by 7.3\% and 5.6\% on Office-31 and Office-Home, respectively.

More recent methods such as POUF~\citep{tanwisuth2023pouf} and ReCLIP~\citep{xuefeng2023reclip} directly adapt CLIP on the target domain without training on the source domain. Although target adaptation can improve the performance of CLIP over the zero-shot version, we note that (1) co-learning can further boost performance by leveraging the alternative classification decisions of the source model, as demonstrated in Table~\ref{tab: office31_poufreclip} and \ref{tab: officehome_poufreclip} on Office-31 and Office-Home, and (2) co-learning with zero-shot CLIP (94.8\%) outperformed both POUF (94.7\%) and ReCLIP (86.2\%) on Office-31. Moreover, the adapted CLIP with ViT-L/14@336 backbone has 428 million parameters, while our co-learned ResNet-50 only has 25 million paramters and can achieve comparable or better performance. In our experiments, since the POUF and ReCLIP classifiers are already adapted to the target task, we directly use these classifiers instead of the weighted NCC in the pre-trained model branch during co-learning.

Moreover, relying solely on CLIP (as in POUF and ReCLIP) is unreliable on object categories that CLIP did not receive sufficient pre-training on. The source model is still necessary in such cases. We experiment with a fine-grained classification dataset with 200 specifies of birds, and compare with POUF which is the stronger of the two methods from Table~\ref{tab: office31_poufreclip} and \ref{tab: officehome_poufreclip}. The source domain is CUB-200-Painting \citep{wang2020cubpainting} containing images of birds in watercolors, oil paintings, pencil drawings, stamps and cartoons. The target domain is CUB-200-2011 \citep{wah2011cub} containing photos of birds. Zero-shot CLIP has 60.7\% accuracy, and POUF has 64.6\% accuracy on the target domain. The Resnet-50 source model has an accuracy of 41.3\%. By making use of both source model and CLIP predictions, \colearnplus\ achieves the highest accuracy of 69.6\%.

\section{Conclusion}
\label{sec: conclusion}

In this work, we explored the use of pre-trained networks beyond its current role as initializations for source training in the source-free domain adaptation (SFDA) pipeline. We observed that source-training can instill source bias and cause large data pre-trained networks to lose their inherent generalization capabilities on the target domain. We introduced an integrated two-branch framework to restore and insert useful target domain information from pre-trained networks during target adaptation. We designed a simple co-learning strategy to collaboratively rectify source bias and enhance target pseudolabel quality to finetune the source model. This flexible framework allows us to integrate modern pre-trained vision and vision-language networks such as transformers and CLIP, thereby leveraging their superior representation learning capabilities. Experimental results on benchmark datasets validate the effectiveness of our proposed framework and strategy.

\backmatter

\bmhead{Acknowledgments}
This research is supported by the Agency for Science, Technology and Research (A*STAR) under its AME Programmatic Funds (Grant No. A20H6b0151).

\bmhead{Data availability statement}
All datasets used in this work are publicly available. Office-31~\citep{Saenko2010AdaptingVC} is available at \url{https://www.cc.gatech.edu/~judy/domainadapt/}. Office-Home~\citep{Venkateswara2017DeepHN} is available at \url{https://www.hemanthdv.org/officeHomeDataset.html}. VisDA-C~\citep{Peng2017VisDATV} is available at \url{https://github.com/VisionLearningGroup/taskcv-2017-public}. DomainNet~\citep{peng2019moment} is available at \url{https://ai.bu.edu/M3SDA/}. CUB-200-2011~\citep{wah2011cub} is available at \url{https://www.vision.caltech.edu/datasets/cub_200_2011/} and CUB-200-Painting is available at \url{https://github.com/thuml/PAN}.

\begin{appendices}

\section{Detailed Results}\label{appendix: detailed results}

In Table~\ref{tab: office31_results_full}, \ref{tab: officehome_results_full} and \ref{tab: visda_results_full}, we provide detailed results for our proposed strategy applied to the datasets Office-31, Office-Home and VisDA-C utilizing the following co-learning networks: ResNet-50, ResNet-101, ConvNeXt-S, Swin-S, ConvNeXt-B, Swin-B and CLIP, where S and B denote the small and base versions of the architectures, respectively. Broadly, in harnessing pre-trained vision models, co-learning with the more recently-released ConvNeXt and Swin networks demonstrates better adaptation performance than co-learning with the ResNets. The introduction of the vision-language CLIP model further enhances co-learning performance compared to the vision models evaluated.
In particular, \colearnplus\ capitalizes on CLIP's text encoder and zero-shot image recognition capabilities, achieving the highest target accuracy in most cases.

In addition, we find that even networks with poor feature extraction ability, such as AlexNet, can contribute useful features to improve performance when the target style closely resembles that of the pre-training dataset. Co-learned with AlexNet, the overall Office-Home accuracy of SHOT and SHOT++ (71.9\% and 72.7\%) is little changed at 71.8\% $\red{(\downarrow 0.1\%)}$ and 72.7\% $\green{(=)}$, respectively. However, 5 out of 12 domain pairs improved by 0.1-1\% and 0.1-1.2\% especially when the target domain is Product or Real World, which has a similar style as the pre-training dataset ImageNet.

\begin{table*}[htb]

\begin{subtable}[t]{\textwidth}
\centering
\begin{adjustbox}{max width=0.7\textwidth}
\begin{tabular}{l*{7}{c}}
\toprule[1pt]\midrule[0.3pt]
\textbf{Method}                 & \multicolumn{7}{c}{\textbf{Office-31}} \\ \cmidrule(lr){2-8}
                                & $\mathbf{A\rightarrow D}$ & $\mathbf{A\rightarrow W}$ & $\mathbf{D\rightarrow A}$  
                                & $\mathbf{D\rightarrow W}$ & $\mathbf{W\rightarrow A}$ & $\mathbf{W\rightarrow D}$ &\textbf{Avg} \\ \midrule
Source Only                     & 81.9  & 78.0  & 59.4  & 93.6  & 63.4  & 98.8  & 79.2 \\
Co-learn (w/ Resnet-50)         & 93.6  & 90.2  & 75.7  & 98.2  & 72.5  & 99.4  & 88.3\\
Co-learn (w/ Resnet-101)        & 94.2  & 91.6  & 74.7  & 98.6  & 75.6  & 99.6  & 89.0 \\
Co-learn (w/ ConvNeXt-S)        & 96.6  & 92.6  & 79.8  & 97.7  & 79.6  & 99.4  & 91.0\\
Co-learn (w/ Swin-S)            & 96.8  & 93.3  & 79.2  & 98.7  & 80.2  & 99.6  & 91.3\\
Co-learn (w/ ConvNeXt-B)        & \textbf{97.8}  & 96.6  & 80.5  & 98.5  & 79.4  & 99.6  & 92.1\\
Co-learn (w/ Swin-B)            & 97.4  & \textbf{98.2}  & \textbf{84.5}  & \underline{\textbf{99.1}}  & \textbf{82.2}  & \textbf{100.0} & \textbf{93.6} \\ \hdashline
Co-learn (w/ CLIP)              & 99.2 & \underline{\textbf{99.7}} & 85.3 & \underline{\textbf{99.1}} & 83.2 & \underline{\textbf{100.0}} & 94.4 \\
\colearnplus (w/ CLIP)          & \underline{\textbf{99.6}} & 99.0 & \underline{\textbf{86.3}} & \underline{\textbf{99.1}} & \underline{\textbf{84.8}} & \underline{\textbf{100.0}} & \underline{\textbf{94.8}}\\
\midrule[0.3pt]\bottomrule[1pt]
\end{tabular}
\end{adjustbox}
\caption{Office-31: 31-class classification accuracy of adapted ResNet-50 \label{tab: office31_results_full}}
\end{subtable}

\begin{subtable}[t]{\textwidth}
\centering
 \setlength{\tabcolsep}{4pt}
\begin{adjustbox}{max width=\textwidth}
\begin{tabular}{l*{13}{c}}
\toprule[1pt]\midrule[0.3pt]
\textbf{Method}                 & \multicolumn{13}{c}{\textbf{Office-Home}} \\ \cmidrule(lr){2-14}
                                & $\mathbf{A\rightarrow C}$ & $\mathbf{A\rightarrow P}$ & $\mathbf{A\rightarrow R}$  
                                & $\mathbf{C\rightarrow A}$ & $\mathbf{C\rightarrow P}$ & $\mathbf{C\rightarrow R}$ 
                                & $\mathbf{P\rightarrow A}$ & $\mathbf{P\rightarrow C}$ & $\mathbf{P\rightarrow R}$
                                & $\mathbf{R\rightarrow A}$ & $\mathbf{R\rightarrow C}$ & $\mathbf{R\rightarrow P}$ & \textbf{Avg}\\ \midrule
Source Only                     & 43.5  & 67.1  & 74.2  & 51.5  & 62.2  & 63.3  & 51.4  & 40.7  & 73.2  & 64.6  & 45.8  & 77.6  & 59.6 \\
Co-learn (w/ Resnet-50)         & 51.8  & 78.9  & 81.3  & 66.7  & 78.8  & 79.4  & 66.3  & 50.0  & 80.6  & 71.1  & 53.7  & 81.3  & 70.0\\
Co-learn (w/ Resnet-101)        & 54.6  & 81.8  & 83.5  & 68.6  & 79.3  & 80.4  & 68.7  & 52.3  & 82.0  & 72.4  & 57.1  & 84.1  & 72.1 \\
Co-learn (w/ ConvNeXt-S)        & 59.7  & 86.3  & 87.1  & 75.9  & 84.5  & 86.8  & 76.1  & 58.7  & 87.1  & 78.0  & 61.9  & 87.2  & 77.4 \\
Co-learn (w/ Swin-S)            & 56.4  & 85.1  & 88.0  & 73.9  & 83.7  & 86.1  & 75.4  & 55.3  & 87.8  & 77.3  & 58.9  & 87.9  & 76.3 \\
Co-learn (w/ ConvNeXt-B)        & 60.5  & 85.9  & 87.2  & 76.1  & 85.3  & 86.6  & 76.5  & 58.6  & 87.5  & 78.9  & 62.4  & 88.8  & 77.9 \\
Co-learn (w/ Swin-B)            & \textbf{69.6}  & \textbf{89.5}  & \textbf{91.2}  & \textbf{82.7}  & \textbf{88.4}  & \textbf{91.3}  & \textbf{82.6}  & \textbf{68.5}  & \textbf{91.5}  & \textbf{82.8}  & \textbf{71.3}  & \textbf{92.1}  & \textbf{83.5} \\ \hdashline
Co-learn (w/ CLIP)              & 77.2 & 90.4 & 91.0 & 77.1 & 88.1 & 90.0 & 76.6 & 72.5 & 90.1 & 82.0 & 79.6 & 93.0 & 84.0 \\
\colearnplus (w/ CLIP)          & \underline{\textbf{80.0}} & \underline{\textbf{91.2}} & \underline{\textbf{91.8}} & \underline{\textbf{83.4}} & \underline{\textbf{92.7}} & \underline{\textbf{91.3}} & \underline{\textbf{83.4}} & \underline{\textbf{78.9}} & \underline{\textbf{92.0}} & \underline{\textbf{85.5}} & \underline{\textbf{80.6}} & \underline{\textbf{94.7}} & \underline{\textbf{87.1}}\\
\midrule[0.3pt]\bottomrule[1pt]
\end{tabular}
\end{adjustbox}
\caption{Office-Home: 65-class classification accuracy of adapted ResNet-50 \label{tab: officehome_results_full}}
\end{subtable}
\end{table*}

\begin{table*}[htb]
\ContinuedFloat
\begin{subtable}[t]{\textwidth}
\centering
\begin{adjustbox}{max width=\textwidth}
\begin{tabular}{l*{13}{c}}
\toprule[1pt]\midrule[0.3pt]
\textbf{Method}                 & \multicolumn{13}{c}{\textbf{VisDA-C}} \\ \cmidrule(lr){2-14}
                                & \textbf{plane} & \textbf{bike} & \textbf{bus} & \textbf{car} & \textbf{horse} & \textbf{knife} & \textbf{mcycle} & \textbf{person} & \textbf{plant} & \textbf{sktbrd} & \textbf{train} & \textbf{truck} & \textbf{Avg}\\ \midrule
Source Only                     & 51.5  & 15.3  & 43.4  & 75.4  & 71.2  & 6.8   & 85.5  & 18.8  & 49.4  & 46.4  & 82.1  & 5.4   & 45.9 \\
Co-learn (w/ Resnet-50)         & 96.2  & 76.2  & 77.5  & 77.8  & 93.8  & 96.6  & 91.5  & 76.7  & 90.4  & 90.8  & 86.0  & 48.9  & 83.5 \\
Co-learn (w/ Resnet-101)        & 96.5  & 78.9  & 77.5  & 75.7  & 94.6  & 95.8  & 89.1  & 77.7  & 90.5  & 91.0  & 86.2  & 51.5  & 83.7 \\
Co-learn (w/ ConvNeXt-S)        & 97.8  & 89.7  & 82.3  & \textbf{81.3}  & 97.3  & 97.8  & 93.4  & 66.9  & 95.4  & 96.0  & 90.7  & \textbf{56.5}  & 87.1 \\
Co-learn (w/ Swin-S)            & 97.8  & 88.5  & 84.7  & 78.5  & 96.8  & 97.8  & 93.3  & 73.9  & 94.9  & 94.8  & 91.2  & 54.8  & 87.2 \\
Co-learn (w/ ConvNeXt-B)        & 98.0  & 89.2  & \textbf{84.9}  & 80.2  & 97.0  & 98.4  & 93.6  & 64.3  & \underline{\textbf{95.6}}  & \textbf{96.3}  & 90.4  & 54.0  & 86.8 \\
Co-learn (w/ Swin-B)            & \textbf{99.0}  & \textbf{90.0}  & 84.2  & 81.0  & \textbf{98.1}  & \textbf{97.9}  & \textbf{94.9}  & \textbf{80.1}  & 94.8  & 95.9  & \textbf{94.4}  & 48.1  & \textbf{88.2} \\ \hdashline
Co-learn (w/ CLIP)              & 98.9 & 93.2 & 81.0 & \underline{\textbf{83.0}} & 98.6 & 98.8 & 95.7 & \underline{\textbf{84.8}} & \textbf{94.8} & 97.3 & 95.1 & 41.6 & 88.6 \\
\colearnplus (w/ CLIP)          & \underline{\textbf{99.6}} & \underline{\textbf{94.6}} & \underline{\textbf{90.9}} & 77.8 & \underline{\textbf{99.6}} & \underline{\textbf{99.0}} & \underline{\textbf{96.4}} & 80.1 & 90.0 & \underline{\textbf{99.2}} & \underline{\textbf{96.3}} & \underline{\textbf{70.1}} & \underline{\textbf{91.1}} \\
\midrule[0.3pt]\bottomrule[1pt]
\end{tabular}
\end{adjustbox}
\caption{VisDA-C: 12-class classification accuracy of adapted ResNet-101 \label{tab: visda_results_full}}
\end{subtable}

\caption{Classification accuracy of adapted models. The source model is initialized with ImageNet-1k ResNet-50 weights for Office-31 and Office-Home, and ImageNet-1k ResNet-101 weights for VisDA-C. For proposed strategy, the pre-trained network used for co-learning is given in parenthesis: CLIP is pre-trained on WIT, and the rest are pre-trained on ImageNet-1k (i.e. no new data is introduced). $\dagger$ denotes reproduced results. \label{tab: results_full}}
\end{table*}

\subsection{Visualization}

\begin{figure*}
    \centering
    \includegraphics[width=\linewidth]{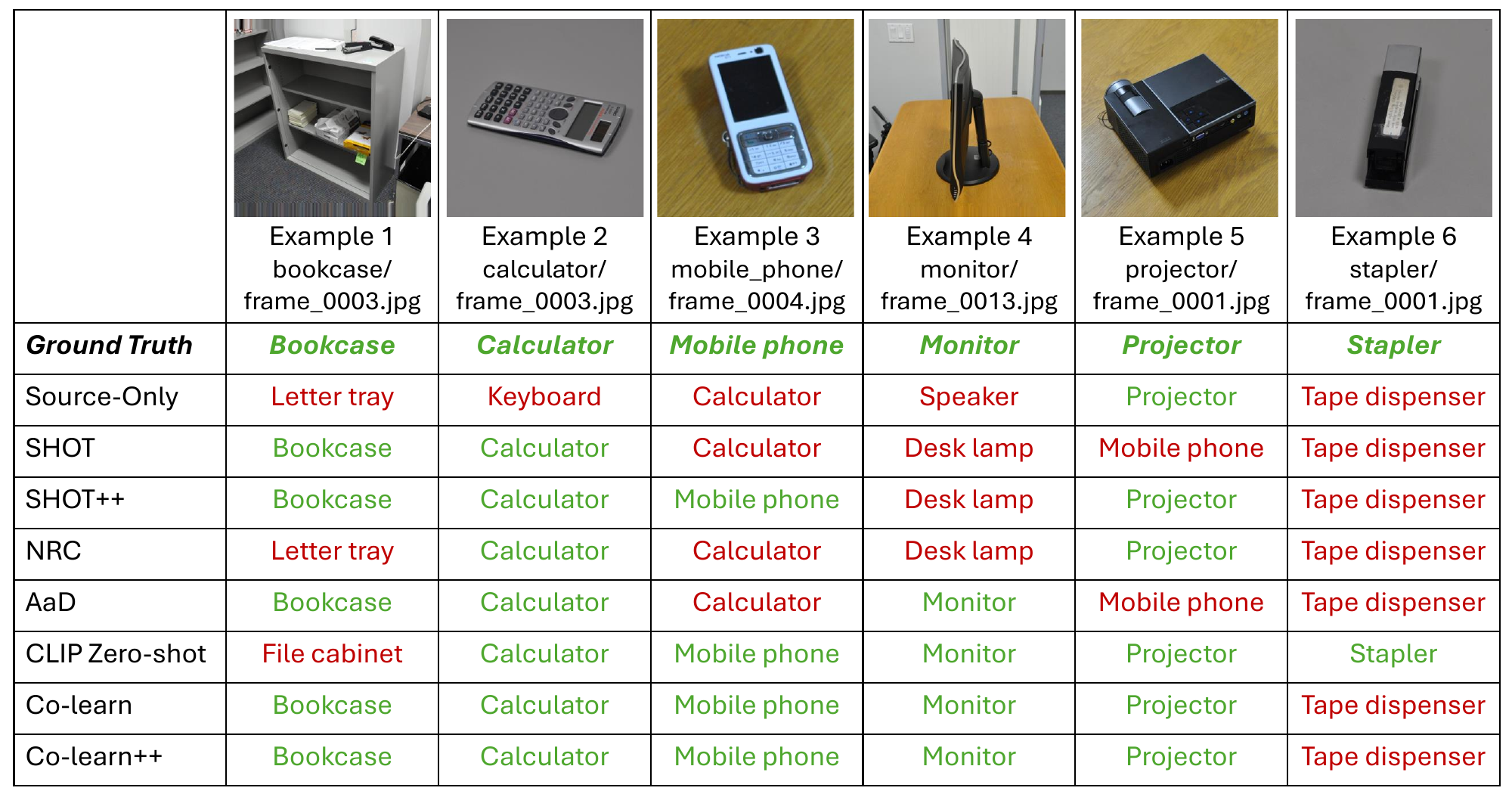}
    \caption{Office-31: Example images and model predictions for $A \rightarrow D$. Co-learning is conducted with CLIP.}
    \label{fig:office_visualize}
\end{figure*}

\begin{figure*}
    \centering
    \includegraphics[width=\linewidth]{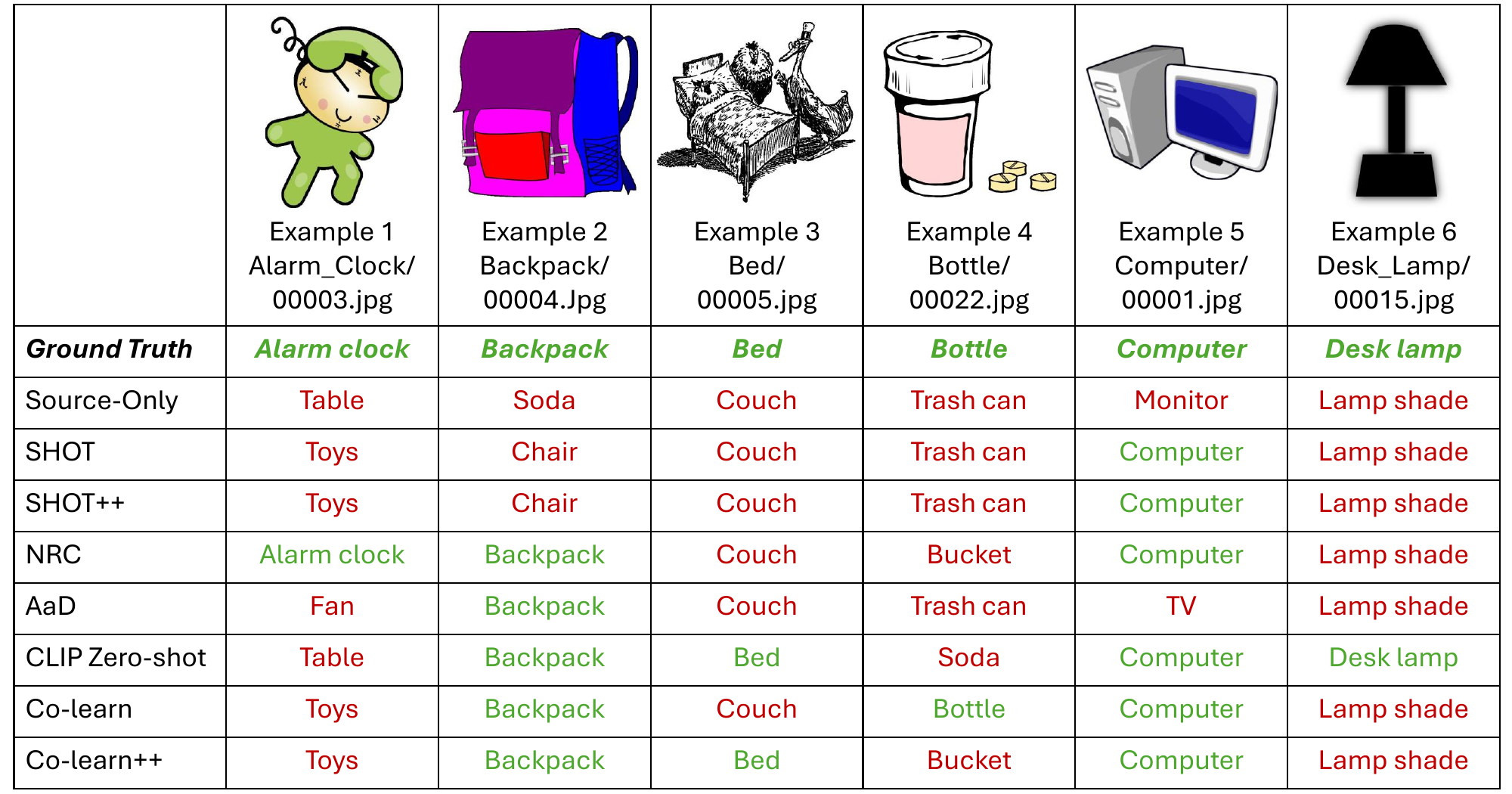}
    \caption{Office-Home: Example images and model predictions for $A \rightarrow C$. Co-learning is conducted with CLIP.}
    \label{fig:officehome_visualize}
\end{figure*}

We provide visualizations of example images and model predictions in Figure~\ref{fig:office_visualize} on Office-31 $A\rightarrow D$ and Figure~\ref{fig:officehome_visualize} on Office-Home $A\rightarrow C$. In particular, we point out some cases where Co-learn and/or \colearnplus\ give the correct predictions following CLIP Zero-shot: Figure~\ref{fig:office_visualize} Example 2, 3, 4, 5 and Figure~\ref{fig:officehome_visualize} Example 2, 3, 5. We note that the co-learned models do not necessarily follow CLIP Zero-shot predictions as learning depends on both the source and pre-trained models, as well as inherent structures in the target dataset.

\section{Integrating zero-shot CLIP}\label{appendix: Integrating zero-shot CLIP}

In Table~\ref{tab: guidance_full}, we provide an expanded version of Table~\ref{tab: clip_guidance} where we list the few-shot target accuracy of CLIP's zero-shot text-classifier and image-classifier for each domain pair. In general, when the image-classifier has better few-shot accuracy than the text-classifier, weak guidance results in better adaptation performance. When the text-classifier has better few-shot accuracy, the strength of guidance from the text-classifier should be increased. There are some exceptions such as in Office-Home $A\rightarrow R$ , $C\rightarrow R$ and $P\rightarrow R$. This is because (i) few-shot sampling is variable and likely does not cover the entire data distribution, and (ii) the few-shot accuracy is evaluated on CLIP and may not correlate exactly with performance of the adapted network. Hence, instead of selecting the guidance strength per domain pair, we take the average few-shot accuracy across domain pairs and select the guidance strength per dataset.

\begin{table*}[htb]

\begin{subtable}[t]{\textwidth}
\centering
\begin{adjustbox}{max width=0.7\textwidth}
\begin{tabular}{l*{7}{c}}
\toprule[1pt]\midrule[0.3pt]
\textbf{Guidance}               & \multicolumn{7}{c}{\textbf{Office-31}} \\ \cmidrule(lr){2-8}
                                & $\mathbf{A\rightarrow D}$ & $\mathbf{A\rightarrow W}$ & $\mathbf{D\rightarrow A}$  
                                & $\mathbf{D\rightarrow W}$ & $\mathbf{W\rightarrow A}$ & $\mathbf{W\rightarrow D}$ &\textbf{Avg} \\ \midrule
weak            & 99.6 & 99.0 & 86.3 & 99.1 & 84.8 & 100.0 & 94.8 \\
strong          & 94.8 & 93.8 & 88.2 & 94.8 & 88.3 & 94.8 & 92.5 \\ \midrule
image-clf@3     & 89.6 & 92.1 & 87.5 & 93.2 & 87.5 & 93.5 & 90.6 \\
text-clf@3      & 86.0 & 88.5 & 88.9 & 87.8 & 88.9 & 90.0 & 88.4 \\
ratio           & $>1$ & $>1$ & $<1$ & $>1$ & $<1$ & $>1$ & $>1$ \\
\midrule[0.3pt]\bottomrule[1pt]
\end{tabular}
\end{adjustbox}
\caption{Office-31: 31-class classification accuracy of adapted ResNet-50 \label{tab: office31_guidance_full}}
\end{subtable}

\begin{subtable}[t]{\textwidth}
\centering
 \setlength{\tabcolsep}{4pt}
\begin{adjustbox}{max width=\textwidth}
\begin{tabular}{l*{13}{c}}
\toprule[1pt]\midrule[0.3pt]
\textbf{Guidance}               & \multicolumn{13}{c}{\textbf{Office-Home}} \\ \cmidrule(lr){2-14}
                                & $\mathbf{A\rightarrow C}$ & $\mathbf{A\rightarrow P}$ & $\mathbf{A\rightarrow R}$  
                                & $\mathbf{C\rightarrow A}$ & $\mathbf{C\rightarrow P}$ & $\mathbf{C\rightarrow R}$ 
                                & $\mathbf{P\rightarrow A}$ & $\mathbf{P\rightarrow C}$ & $\mathbf{P\rightarrow R}$
                                & $\mathbf{R\rightarrow A}$ & $\mathbf{R\rightarrow C}$ & $\mathbf{R\rightarrow P}$ & \textbf{Avg}\\ \midrule
weak            & 80.0 & 91.2 & 91.8 & 83.4 & 92.7 & 91.3 & 83.4 & 78.9 & 92.0 & 85.5 & 80.6 & 94.7 & 87.1 \\
strong          & 76.4 & 88.5 & 85.8 & 84.2 & 88.6 & 85.7 & 84.3 & 75.7 & 85.7 & 84.8 & 75.8 & 88.6 & 83.7 \\ \midrule
image-clf@3     & 72.3 & 84.1 & 85.3 & 81.0 & 86.5 & 85.8 & 81.0 & 74.0 & 82.4 & 81.0 & 74.0 & 88.0 & 81.3 \\
text-clf@3      & 69.9 & 83.6 & 86.3 & 82.1 & 86.5 & 86.0 & 82.1 & 71.3 & 83.9 & 82.1 & 71.3 & 86.8 & 81.0 \\
ratio           & $>1$ & $>1$ & $<1$ & $<1$ & $=1$ & $<1$ & $<1$ & $>1$ & $<1$ & $<1$ & $>1$ & $>1$ & $>1$ \\
\midrule[0.3pt]\bottomrule[1pt]
\end{tabular}
\end{adjustbox}
\caption{Office-Home: 65-class classification accuracy of adapted ResNet-50 \label{tab: officehome_guidance_full}}
\end{subtable}
\end{table*}

\begin{table*}[htb]
\ContinuedFloat
\begin{subtable}[t]{\textwidth}
\centering
\begin{adjustbox}{max width=\textwidth}
\begin{tabular}{l*{13}{c}}
\toprule[1pt]\midrule[0.3pt]
\textbf{Guidance}               & \multicolumn{13}{c}{\textbf{DomainNet}} \\ \cmidrule(lr){2-14}
                                & $\mathbf{C\rightarrow P}$ & $\mathbf{C\rightarrow R}$ & $\mathbf{C\rightarrow S}$  
                                & $\mathbf{P\rightarrow C}$ & $\mathbf{P\rightarrow R}$ & $\mathbf{P\rightarrow S}$ 
                                & $\mathbf{R\rightarrow C}$ & $\mathbf{R\rightarrow P}$ & $\mathbf{R\rightarrow S}$
                                & $\mathbf{S\rightarrow C}$ & $\mathbf{S\rightarrow P}$ & $\mathbf{S\rightarrow R}$ & \textbf{Avg}\\ \midrule
weak            & 83.0 & 90.1 & 84.7 & 88.1 & 90.8 & 84.6 & 89.6 & 85.0 & 84.2 & 88.3 & 83.3 & 90.2 & 86.8 \\
strong          & 89.5 & 93.9 & 88.6 & 90.0 & 93.8 & 88.7 & 90.3 & 89.4 & 88.5 & 90.1 & 89.5 & 93.9 & 90.5 \\ \midrule
image-clf@3     & 86.3 & 92.2 & 86.7 & 89.1 & 92.9 & 84.6 & 89.1 & 87.0 & 85.2 & 89.1 & 87.0 & 92.8 & 88.5 \\
text-clf@3      & 90.4 & 94.0 & 88.4 & 89.1 & 94.4 & 86.2 & 89.1 & 89.5 & 88.3 & 89.1 & 89.5 & 95.1 & 90.2 \\
ratio           & $<1$ & $<1$ & $<1$ & $=1$ & $<1$ & $<1$ & $=1$ & $<1$ & $<1$ & $=1$ & $<1$ & $<1$ & $<1$\\
\midrule[0.3pt]\bottomrule[1pt]
\end{tabular}
\end{adjustbox}
\caption{DomainNet: 126-class classification accuracy of adapted ResNet-50 \label{tab: domainnet_guidance_full}}
\end{subtable}

\caption{Comparison of \colearnplus\ classification accuracy with different strength of CLIP's text-classifier-based zero-shot guidance. The values `image-clf@3' and `text-clf@3' measure the 3-shot target domain accuracy of the image-classifier and text-classifier, respectively. \label{tab: guidance_full}}
\end{table*}

\end{appendices}

\bibliography{sn-bibliography}

\end{document}